%% file: main.tex
\documentclass{article}
\input{common}

\newtheorem{theorem}{Theorem}[section]

\newcommand{\balgorithm}  {\begin{algorithm}}
\newcommand{\ealgorithm}  {\end{algorithm}}
\newcommand{\balgorithmic}{\begin{algorithmic}}
\newcommand{\ealgorithmic}{\end{algorithmic}}

    \usepackage[preprint]{neurips_2024}

\usepackage[utf8]{inputenc} 
\usepackage[T1]{fontenc}    
\usepackage{hyperref}       
\usepackage{url}            
\usepackage{booktabs}       
\usepackage{amsfonts}       
\usepackage{nicefrac}       
\usepackage{microtype}      
\usepackage{xcolor}         

\usepackage{amsmath}
\usepackage{graphicx}
\usepackage{wrapfig}
\usepackage{bbding}
\usepackage{makecell}
\usepackage{wrapfig}
\usepackage[numbers]{natbib}
\usepackage{multirow}
\usepackage{booktabs}

\title{S-STE: Continuous Pruning Function for \\Efficient 2:4 Sparse Pre-training}

\author{%
  Yuezhou Hu$^{1}$,~~Jun Zhu$^{1}$,~~Jianfei Chen$^{1\dag}$ \\
  $^1$Dept. of Comp. Sci. \& Tech., Institute for AI, BNRist Center, \\Tsinghua-Bosch Joint ML Center, THBI Lab, Tsinghua University. \\
  \texttt{huyz21@mails.tsinghua.edu.cn, \{dcszj,jianfeic\}@tsinghua.edu.cn} \\
}

\begin{document}

\maketitle

\def\customfootnotetext#1#2{{%
  \let\thefootnote\relax
  \footnotetext[#1]{#2}}}

\customfootnotetext{1}{\textsuperscript{\dag}Corresponding author.}

\begin{abstract}
Training deep neural networks (DNNs) is costly. Fortunately, Nvidia Ampere and Hopper GPUs can accelerate matrix multiplications twice as fast as a dense equivalent by implementing 2:4 sparsity. However, previous STE-based 2:4 pre-training methods (\eg~STE with hard-thresholding, SR-STE) suffer from optimization difficulties because of discontinuous pruning function.
In this study, we comprehensively analyse the bottleneck of traditional N:M sparse training and recognize three drawbacks with discontinuity: incorrect descending direction, inability to predict the amount of descent and sparse mask oscillation. In light of this, we propose S-STE, a simple yet powerful 2:4 training method that contains two parts: to continuously project weights to be 2:4 sparse, and to rescale sparse weights with a per-tensor fixed scaling factor. Besides, we adopt minimum-variance unbiased estimation for activation gradient and FP8 quantization for whole process. Results show that our method surpasses previous 2:4 pre-training recipes and is comparable even with full parameter models. Our toolkit is available at \url{https://github.com/huyz2023/2by4-pretrain}.
\end{abstract}
\input{1-intro}

\input{2-formulation}
\input{3-discontinuity}
\input{4-method}

\input{5-implementation}

\input{6-experiments}

\input{7-related}
\input{8-conclusion}

\begin{ack}
We would like to thank Ziteng Wang, Bingrui Li, Haocheng Xi, Kang Zhao and Jintao Zhang, Brian Chmiel and Daniel Soudry for valuable discussions. This work was supported by the National Key Research and Development Program of China (No.~2021ZD0110502) and NSFC Project (Nos.~62376131). J.Z is
also supported by the XPlorer Prize.
\end{ack}

{
\small
\bibliographystyle{plainnat}
\bibliography{reference}
}
\newpage
\appendix
\input{appendix}

\end{document}

%% file: common.tex
\usepackage{color}
\usepackage{bm}
\usepackage{hyperref}
\usepackage{amsmath,amsfonts,amsthm}
\usepackage{blindtext}
\usepackage{listings}
\usepackage{algorithm, algorithmic}
\usepackage{booktabs}
\usepackage{multirow}
\usepackage{multicol}
\usepackage{caption}
\usepackage{enumitem} 

\newcommand{\ie}{\emph{i.e.}}
\newcommand{\eg}{\emph{e.g.}}
\newcommand{\st}{\emph{s.t.}}

\newlist{myitems}{enumerate}{3}
\setlist[myitems, 1]
{label=\arabic{myitemsi}.,leftmargin=15pt,labelwidth=10pt,labelsep=5pt,
topsep=0pt,parsep=0pt,partopsep=0pt,noitemsep
}

\usepackage{ifthen}

\newif\ifdebug
\debugfalse

\newcommand{\jianfei}[1]{{\color{red}{\bf\sf [Jianfei: #1]}}}

\newcommand{\vect}[1]{\boldsymbol{\mathbf{#1}}}

\newcommand{\av}{\vect a}

\newcommand{\mv}{\vect m}

\newcommand{\wv}{\vect w}

\newcommand{\Iv}{\vect I}

\newcommand{\Wv}{\vect W}
\newcommand{\Xv}{\vect X}

\newcommand{\Zv}{\vect Z}

\newcommand{\Dc}{\mathcal D}

\newcommand{\Mc}{\mathcal M}

\newcommand{\Wc}{\mathcal W}

\newcommand{\Rb}{\mathbb R}

\newcommand{\norm}[1]{\left\lVert#1\right\rVert}
\newcommand{\abs}[1]{\left|#1\right|}

\makeatletter
\newtheorem*{rep@theorem}{\rep@title}
\newcommand{\newreptheorem}[2]{%
	\newenvironment{rep#1}[1]{%
		\def\rep@title{#2 \ref{##1}}%
		\begin{rep@theorem}}%
		{\end{rep@theorem}}}
\makeatother

%% file: 1-intro.tex
\section{Introduction}

Large scale transformers have achieved many impressive results such as chatbots~\cite{touvron2023llama}, text-to-video generation~\cite{liu2024sora}, and robot manipulation~\cite{brohan2023rt2}. However, the pre-training of these models is extremely expensive, typically requiring thousands of GPUs to train for months~\cite{brown2020language}.
One possible way to accelerate deep learning computation is sparsity. N:M sparsity \cite{mishra2021accelerating} is a hardware-friendly sparsity pattern, where every group of $M$ dimensions only has $N$ non-zero entries. Nvidia Ampere GPUs can multiply a 2:4 sparse matrix with a dense matrix, twice as fast as multiplying two dense matrices.

While N:M sparsity has been successfully applied to accelerate inference~\cite{mishra2021accelerating,sun2023wanda,frantar2023sparsegpt,Pandey2007RIAAR,dathathri2020plug}, extending the acceleration to pre-training is highly challenging. To accelerate pre-training, the sparse model must be trained from scratch (random initialization), and the network must stay sparse at all training iterations. To meet these requirements, the algorithm should be able to actively explore connectivity patterns within the constrained N:M parameter space. Therefore, popular pruning methods such as single-shot pruning~\cite{lee2019snip}, iterative magnitude pruning~\cite{frankle2020linear,maene2021understanding}, and RigL~\cite{evci2021rigging} cannot be directly applied to this scenario. Moreover, besides forward propagation, the matrix multiplications in back propagation must be sparsified as well, to provide reasonable training speedup. 

Methods based on the straight-through estimator (STE)~\cite{zhou2021learning,bengio2013estimating} have shown promise towards solving the challenging problem of sparse pre-training. They maintain a dense weight, which is sparsified in each iteration for fast forward\&backward computation, and the dense weight is then updated with STE gradients. In this way, connectivity patterns can be learned jointly with weights in an end-to-end fashion with stochastic gradient optimizers. 
SR-STE~\cite{zhou2021learning} is such a method to train sparse networks from scratch, with a regularization term to stabilize the training. Several subsequent works~\cite{hubara2021accelerated,zhang2023bidirectional,chmiel2023minimum} further accelerate back propagation with sparse computations, and ~\citet{hu2024accelerating} applied it for pre-training language models. 
However, these sparse training methods still have an accuracy gap compared to dense training. Moreover, SR-STE introduces a regularization strength hyper-parameter, which is hard to tune. Due to these limitations, N:M sparsity is not yet used to accelerate pre-training. 

In this work, we study STE-based pre-training from the optimization perspective. We point out that STE-based pre-training defines a \emph{discontinuous} loss function, which existing optimization theory and algorithms cannot handle. We reveal several intriguing phenomena highlighting the difficulty of discontinuous optimization, including incorrect descending direction, inability to predict the amount of descent, and oscillation. 
We sidestep the curse of discontinuity by proposing smooth straight-through estimator (S-STE) as a solution. Cruically, S-STE introduces a new pruning function, which uses a continuous projection function to prune weights to be 2:4 sparse, and scales all nonzero elements to minimize the mean-square-error between original dense weight vector and sparse weight vector. The proposed 2:4 soft-thresholding function is \emph{continuous} but can still generate N:M sparse weights at all times. In this way, the objective function is continuous, and gradient-based optimizers can be readily used. Furthermore, S-STE does not introduce any hyper-parameter, so its practical adoption is easier than SR-STE. 

We devise comprehensive pre-training experiments on S-STE, including WMT machine translation, GPT-2 pre-training and, DeiT image classification. Results show that our method surpass previous 2:4 pre-training recipes on a wide range of tasks.

%% file: 2-formulation.tex
\section{Formulation of sparse pre-training}
\label{sec:2}
The training a neural network can be formalized as an optimization problem
$
    \min_{\wv} F(\wv),
$
where $\wv \in \mathbb{R}^d$ is the parameter and $F$ is a differentiable empirical risk function: $
    F(\wv) = R_n(\wv) = \frac{1}{n} \sum \limits_{i=1}^n f (\wv;\xi_{[i]}).
$
Here, $f$ is the loss function, $n$ is the size of data set $\Dc = \{\xi_{[i]}\}_{i=1}^n$ and $\xi_{[i]}$ is the $i$-th sample. 
The optimization can be solved with standard stochastic gradient
method (SG) \cite{bottou2018optimization}. Suppose the network is initialized with $\wv_1$, $\{\alpha_k\}$ is a positive learning rate sequence, and $\xi_{[i_k]}$ is randomly chosen from $\{\xi_{[i]}\}_{i=1}^n$. Then, iteratively we have
$
    \wv_{k+1} = \wv_{k}-\alpha_k \nabla_{\wv_k} f({\wv_k};\xi_{[i_k]}).
$
As we consider pre-training tasks, $\wv_1$ is simply a random initialization. 

The training of a sparse network involves optimizing the parameter $\wv$ in a constrained space $\Wc\subset \Rb^d$. For an N:M-sparse network, the parameter can only have $N$ non-zero elements in each contiguous $M$ dimensions.

Alternative to constrained optimization, we can solve the unconstrained problem:
\begin{align}\label{eqn:sparse-training}
\min_{\wv} F(\tilde \wv) \mbox{ where } \tilde \wv = S(\wv).
\end{align}
Here, $S$ is a pruning function which converts a dense weight $\wv$ to a sparse weight $\tilde \wv\in \Wc$. One common choice is the hard-thresholding pruning function \cite{zhou2021learning, vanderschueren2022straightthrough}. 
For every block of four adjacent elements $ \av = [a_1,...,a_M]^\top \in \mathbb{R}^M$ in the weight vector $\wv$, the pruning function can be defined as
\begin{equation}\label{eqn:hard-thresholding}
    (S_h(\av))_i =
    \begin{cases}
    a_i& \text{ if } |a_i| \geq t\\
    0& \text{ if } |a_i| < t
    \end{cases}, \text{ for }i=1,...,M,
\end{equation}
where $t$ is $N$-th largest element in $\av$.\footnote{In this paper, when talking about large and small, we refer to the magnitude. For example, ``second largest element'' means the element with second largest absolute value.} This essentially performs magnitude-based pruning, by zeroing out the two smallest elements. The hard thresholding function can also be written as $S_h(\av) = \av \odot m_h(\av)$, where $m_h(\av)$ is a 0/1 mask vector, with $(m_h(\av))_i = 1$ if $\abs{a_i}>t$.


However, Eq.~(\ref{eqn:sparse-training}) cannot be directly optimized since the pruning function $S$ is not differentiable. Particularly, the derivative of the hard-thresholding function $S_h$ is undefined on boundary where the second largest and third largest element have the same magnitude. Therefore, straight-through estimator (STE) \cite{zhou2021learning} is for training, by approximating $\nabla_{\wv} f\approx \nabla_{\tilde \wv} f$ and therefore $\partial S_h(\av) / \partial \av \approx \Iv$:
\begin{equation}
\label{eq:7}
    \wv_{k+1} = \wv_{k}-\alpha_k \nabla_{\tilde \wv_k} f(\tilde \wv_k;\xi_{[i_k]}).
\end{equation}

With the pruning function and STE, each iteration of sparse training involves: (1) prune the dense weight  to get the sparse weight: $\tilde \wv = S(\wv)$; (2) compute the loss and gradient with the \emph{sparse} weight; and (3) update the \emph{dense} weight with the gradient. 
Among these, step 2 is most time-consuming, and it can be accelerated with sparse tensor cores given $\tilde \wv$ is N:M-sparse. Next, we will focus on the optimization aspects of sparse training and defer the discussion of computation details to Sec.~\ref{sec:method}.



%% file: 3-discontinuity.tex
\section{The curse of discontinuity}
\label{sec:3}

Classical stochastic optimization theory~\cite{bottou2018optimization} guarantees the convergence for nonconvex and \emph{smooth} (i.e., differentiable with Lipschitz continuous gradients) objective $F$. It can be also extended to handle non-differentiable functions such as ReLU~\cite{li2017convergence}. The real problem of STE-based sparse training is the \emph{discontinuity} of the pruning function $S_h$, as visualized in Fig.~\ref{fig:2}. For a discontinuous function, an arbitrarily small change 
in input $\av$ can cause an unbounded change of the output $S_h(\av)$. 
Such discontinuity appears on the boundary when the $N$-th and $N+1$-th largest elements have same magnitude. 
For example, for a 1:2-sparse pruning function, $S_h(1, 0.999)=(1, 0)$, but $S_h(0.999, 1)=(0, 1)$, and the boundary is the line $a_1 = a_2$. 

\begin{wrapfigure}{r}{0.5\textwidth}
\includegraphics[width=0.4\textwidth]{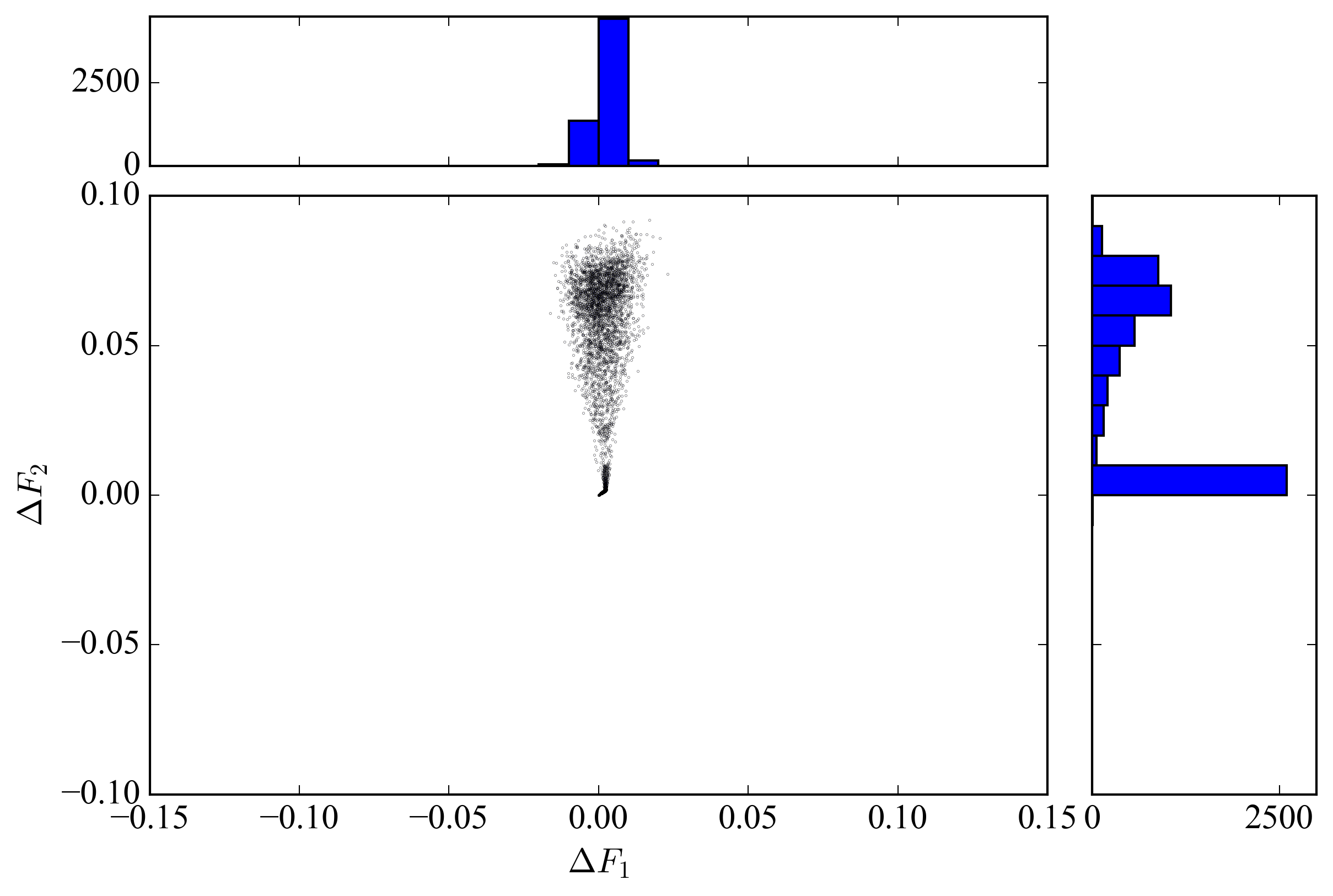}
\caption{Scatter plot of $\Delta F_1$ with $\Delta F_2$ and their distributions on GPT-2 small 124M for iteration $k\in [1,6000]$.}
    \label{fig:0}
\end{wrapfigure}

When $S_h$ is discontinuous, the parameter space $\Rb^d$ can be partitioned into regions $\{\Wc_{\mv}  | \mv \in \Mc\}$, where $\Mc\subset \{0, 1\}^d$ is the space of 0/1 masks with N:M pattern,  and all the parameters  in each region $\wv\in \Wc_{\mv}$ have the same mask $m_h(\wv)=\mv$. The loss landscape $F(S_h(\wv)) = F(m_h(\wv)\odot \wv)$ is continuous and differentiable within each region, where gradient-based algorithms can work well. However, when the optimization trajectory crosses the boundary: $m_h(\wv_{k+1})\ne m_h(\wv_k)$, the behavior is unpredictable. We highlight several intriguing phenomena observed in optimizing such discontinuous objective. We study these phenomena in both a toy problem and real neural networks.

%

\subsection{Phenomenon 1: incorrect descending direction}\label{sec:difficulty-1}




Here, we run a gradient descent algorithm (without stochasticity)  on a small dataset. For a dense model where $F$ is differentiable, with Taylor's formula we should have 
\begin{align}\label{eqn:taylors}
F(\wv_k) - F(\wv_{k+1})\approx (\nabla_{\wv_k} F(\wv_k))^\top (\wv_k - \wv_{k+1}) = \alpha_k \norm{\nabla_{\wv_k} F(\wv_k)}^2 \ge 0.
\end{align}
That is, the objective function will monotonically decrease in each iteration once the learning rate $\alpha_k$ is sufficiently small. However, it is not the case for sparse training. In Fig.~\ref{fig:2}(d), we measure the distribution of the amount of descent (AoD) $\Delta F_k := F(\wv_k) - F(\wv_{k+1})$ for training a GPT-2 large 774M model with Eq.~(\ref{eqn:hard-thresholding},~\ref{eq:7}), across each iteration $k$. The results clearly shows that the objective frequently fails to descent.

We can take a closer look to the weight and mask sequence $(\wv_k, m_h(\wv_k))$ generated by the training algorithm. We compare the following two quantities: the AoD by updating both weight and mask $\Delta F_1 = F(\wv_{k} \odot \mv_{k})-F(\wv_{k+1} \odot \mv_{k+1})$ and the AoD by only updating the weight $\Delta F_2 = F(\wv_{k} \odot \mv_{k})-F(\wv_{k+1} \odot \mv_{k})$. In Fig.~\ref{fig:0}, we can observe $\Delta F_2$ is mostly positive due to the piece-wise continuity of $F$. However, $\Delta F_1$ is frequently negative and very often even smaller than $\Delta F_2$ (updating mask is worse than not updating). This indicates that the main problem is the discontinuity make it hard to estimate the correct descending direction of $\mv$.




\subsection{Phenomenon 2: inability to predict the amount of descent}

\begin{figure}[h]
    \centering
    \includegraphics[width=0.9\linewidth]{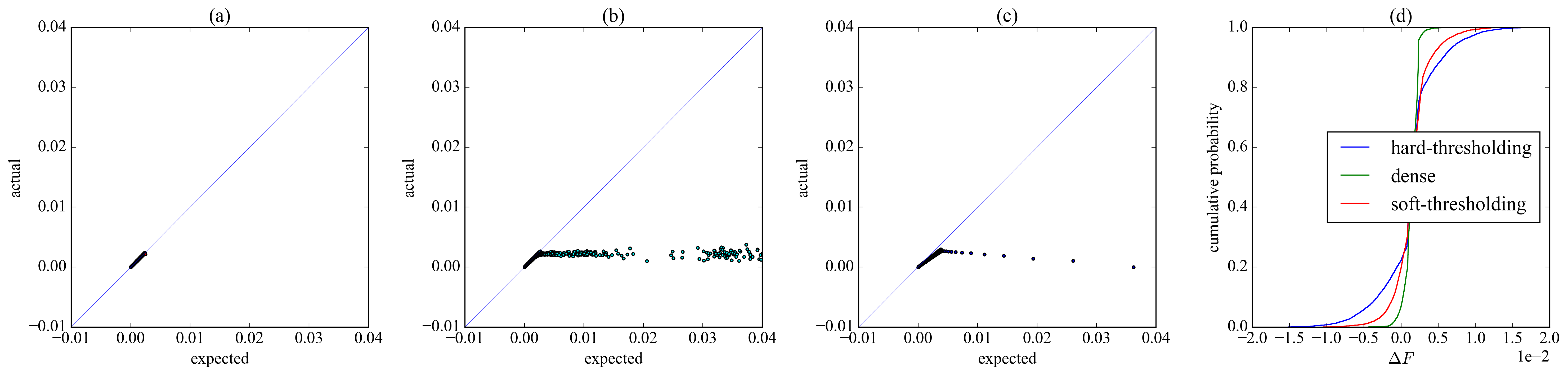}
    \caption{(a)-(c) shows scatter plots of the predicted and actual loss reduction of dense, hard-thresholding and S-STE with GPT-2 large 774M model for iteration $k\in [1,3000]$. The diagonal line is for reference. (d) shows empirical cumulative distribution of their actual AoD for $k\in [1,6000]$.}
    \label{fig:2}
\end{figure}

Besides making mistakes in finding the correct descending direction, algorithms do not know that they make a mistake, in the sense that they fail to predict the AoD at each step. From Eq.~(\ref{eqn:taylors}), we should have $F(\wv_k) - F(\wv_{k+1})\approx (\nabla_{\wv_k} F(\wv_k))^\top (\wv_k - \wv_{k+1})$, where the left hand side is the \emph{actual} AoD, and the right hand side is the \emph{predicted} AoD. We plot the actual AoD against predicted AoD for dense (Fig.~\ref{fig:2}(a)) and sparse training (Fig.~\ref{fig:2}(b)). While for dense training, the two quantities closely matches, for hard-thresholding the actual AoD is often lower for the predicted AoD, particularly when the predicted AoD is large.
To understand this, note that Eq.~(\ref{eqn:taylors}) only holds for $\wv\in \Wc_{m_h(\wv)}$.  Once $\wv_{k+1}-\wv_k$ is large enough that $m_h(\wv_{k+1})\ne m_h(\wv_k)$, the function crosses a border of $S_h$, and $F$ will have a sudden change which is unpredictable by the gradient.  

\subsection{Phenomenon 3: oscillation}
Oscillation is probably the most significant problem in STE-based sparse training. Here, we revisit existing discussions about oscillation~\cite{zhou2021learning,lu2023step,hu2024accelerating}, and then illustrate this issue using a toy example. 

\paragraph{Flip rate}
Flip rate is a simple metric to measure the stability of sparse training \cite{hu2024accelerating}:
$
    r_{k} = \norm{m_h(\wv_k)\oplus m_h(\wv_{k-1})}_1/d,$
where $\oplus$ indicates XOR operation. As \citet{hu2024accelerating} points out, taking the flip rate of the dense model as standard, they observe larger flip rate of hard-thresholding: when training transformers, the flip rate can stay at 6\% in the entire training process. However, a healthy training process should have a large flip rate in the early stage to explore connectivity patterns, and the flip rate should decrease to zero in later stage for the optimization to converge. \citet{hu2024accelerating} describe this phenomenon as ``flip rate explosion'', which is harmful to sparse training.

\paragraph{An exemplar toy problem}
Modern deep neuron networks have billions of parameters and is not strictly convex. These non-ideal conditions make our analysis more difficult with sparse weights. To analyze the characteristics of STE and hard-thresholding on the smallest problem, we devise a simple toy problem that contains two parameters: $\min_{w_1,w_2} g(w_1,w_2) = (w_1-w_2)^2$. This may differ from the real DNN optimization problem, but can help us understand what happens in the process. We are going to show that while using a feasible $\alpha_k$ that can make the dense model converge to global minima, STE with hard-thresholding fails to converge and it oscillates back and forth.

First, for the dense model, the global minima lies on the line $w_1=w_2$. Suppose we start from $\wv_{1} =[0.2,0.1]^\top$, by taking $\alpha_k=0.25$ we can reach global minima in one step. On the other hand, if we are in 1:2 sparse situation, the global minima should be the point $w_1=w_2=0$. By starting from $\wv_{1} =[0.2,0.1]^\top$ and taking $\alpha_k=0.25$, we invariably jumps between $\wv_{2t+1}=[0.2,0.1]^\top$ and $\wv_{2t} = [0.1,0.2]^\top$, and $g$ never decreases.


High flip rate is harmful, because there are frequent changes on the connection of neurons, which means that a number of previous optimization steps on the neuron is deprecated. That is fatal at the end of training \cite{hu2024accelerating}.
The reason of high flip rate on hard-thresholding can be explained by discontinuity: as there are no gentle transitions on both sides of the border, the gradient on the boundary is inaccurate and is unable to indicate the right descending direction. This misalignment is easy to make the tuple $\av$ to oscillate back and forth near the boundary, and cause extremely higher flip rate than the dense model.

\subsection{Overcoming the curse of discontinuity}
 
One way to mitigate discontinuity is sparse-reﬁned straight-through estimator (SR-STE), which adds a sparse-reﬁned regularization on the gradients \cite{zhou2021learning}: $
\min_{\mathbf{w}} F(\mathbf{\tilde w})+\tfrac{\lambda_W}{2} \Vert \wv \odot \overline{m(\wv)}\Vert_2^2
$. While SR-STE works on a wide range of tasks and optimization algorithms \cite{zhou2021learning, hu2024accelerating}, it still has some issues. First, the performance is quite sensitive to the hyper-parameter $\lambda_W$. Second, the new regularization term leads to a competition between loss function and sparse regularization. Finally, the loss function is still discontinuous unless $\lambda_W\rightarrow \infty$.

From the above analysis, discontinuity causes optimization problems. It would be ideal to have a \emph{continuous} pruning function, yet the iterates $(\tilde \wv_k)$ still need to be sparse during the entire training process.

\begin{figure}[ht]
    \centering
    \includegraphics[width=0.9\linewidth]{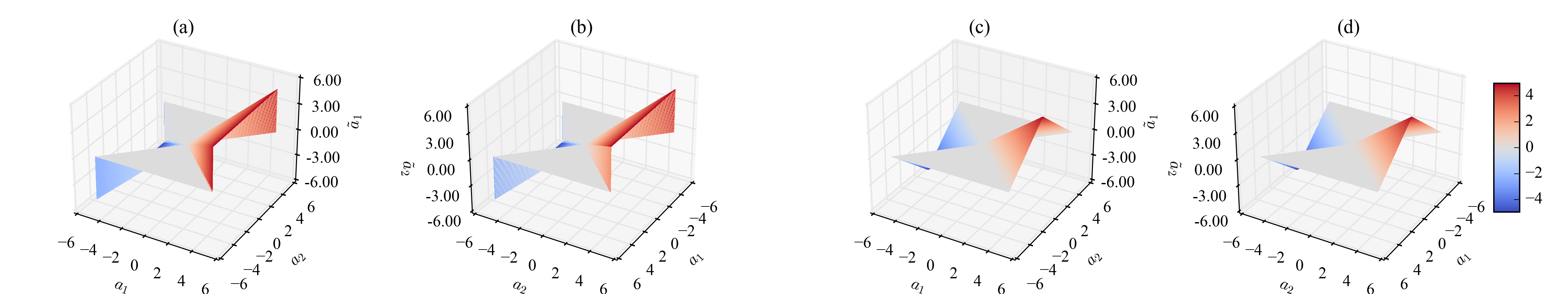}
    \caption{Pruning function of hard-thresholding and soft-thresholding for 1:2-sparsity. (a)(b) show the outputs of hard-thresholding, and (c)(d) show that of soft-thresholding. A sudden jump exists in hard-thresholding if $|a_1|=|a_2|$, while soft-thresholding is continuous in the domain.}
    \label{fig:1}
\end{figure}

%% file: 4-method.tex
\section{Methodology}
\label{sec:method}

\begin{wraptable}{r}{6cm}
\centering
\caption{Validation loss and test accuracy of S-STE with different $\gamma$ on Transformer-base.}
\label{table:3}
\begin{center}
\begin{tabular}{lll}
\toprule
$\gamma$    & Val loss & Test BLEU \\
    \midrule
0   & 4.007           & 26.30      \\
0.33 & 4.014           & 26.01     \\
0.67 & 4.015           & 26.16     \\
1   & 4.072           & 25.63    \\
\bottomrule
\end{tabular}
\end{center}
\vskip -0.1in
\end{wraptable}

In this section we propose a training algorithm (smooth straight-through estimator, S-STE) that contains two main parts combined with STE: 2:4 specific soft-thresholding, and fixed weight rescaling
. They together work as the sparsifying function described in Sec. \ref{sec:2}: $\tilde\wv = S(\wv) = \beta S_{soft}(\wv)$. Results in Fig. \ref{fig:2}(c)(d) and \ref{fig:3456}(d) show that S-STE successfully overcome the three curses of discontinuity. Notably, flip rate curves of S-STE are surprisingly consistent with their dense counterparts, indicating that S-STE is more natural and feasible than SR-STE.

\subsection{2:4 specific soft-thresholding \texorpdfstring{$S_{soft}$}{ssoft}}
\paragraph{Motivation for the design}

As discussed in Sec. \ref{sec:3}, hard-thresholding suffer from the discontinuous problem near the boundary of taking a flip. When input vector changes continuously across the border, two of the four elements simultaneously jump between zeroes and none-zero values. In a continuous pruning function, we need to overcome this drawback and keep these two elements zero on both sides of the border. This means when a flip happens in a four-element block, at lease three of the target elements should be zeroed out simultaneously (except the largest one).

With the above analysis, we modify soft-thresholding function for traditional pruning in \citet{vanderschueren2022straightthrough} as our 2:4 specific soft-thresholding. Given a vector $ \av = [a_1, a_2, a_3, a_4]^\top \in \mathbb{R}^4$, S-STE picks out the largest two elements and meanwhile, subtracts the third largest element from weight magnitudes. Assume, without loss of generality, that $[t_1,t_2,t_3,t_4]$ is an rearrangement of $\av^\top$, \st $|t_1|\leq |t_2| \leq |t_3| \leq |t_4|$. Then, the pruning function can be defined as
\begin{equation}
    (S_{soft}(\av))_i = \begin{cases}
        a_i-t& \text{ if } a_i \in [t, +\infty) \\
        0& \text{ if } a_i \in (-t, t) \\
        a_i+t& \text{ if } a_i \in (-\infty, -t]
    \end{cases}, \text{ where } t = |t_2|.
\end{equation}
The plots of soft-thresholding is drawn in Fig. \ref{fig:1}, showing $S_{soft}$ is continuous everywhere. Note that although we define $S_{soft}$ by a block $\av \in \mathbb{R}^4$, $S_{soft}$ can be extended to arbitrary $\av \in \mathbb{R}^{4t}$ for $t\geq 1$, by doing block-wise pruning.

\begin{theorem}
\label{thm:1}
$S_{soft}(\av)$ is a continuous projection for $\av \in \mathbb{R}^d$.
\end{theorem}
A detailed discussion of the proof can be found in Appendix \ref{proof:1}.

\paragraph{Choosing optimal threshold}
Theoretically, any real number in $[|t_2|, |t_3|]$ can be used as a feasible threshold. This gives us infinite options and we describe it with an interpolation as $t = \gamma |t_2| + (1-\gamma) |t_3|$ with $\gamma \in [0,1]$. The larger $\gamma$ is, the closer $t$ is to $|t_3|$, and the smaller $\norm{S_{soft}(\av)}$ is. This may affect model's capacity. In order to maximize the retention of information, using a small $\gamma$ is necessary. In our method we propose to set $\gamma=0$. Experimental results in Table \ref{table:3} also show that the network has the best accuracy when $\gamma = 0$, \ie, $t=|t_2|$.

\subsection{Fixed weight rescaling \texorpdfstring{$\beta$}{beta}}

Because $S_{soft}$ reduce the total magnitude of $\wv$, it is not a close simulation of dense weights. Like \citet{vanderschueren2022straightthrough}, we scale up $S_{soft}(\wv)$ in our method as $S(\wv) = \beta S_{soft}(\wv)$, but we modify weight rescaling in their study to adapt to our approach. First, we use a \emph{per-tensor scale} $\beta$ rather than a per-channel $\beta$ for simplicity. Besides, two important improvements are made: to compute scale factor only at the beginning of training, rather than to dynamically update scale factor during training, and to minimize the mean-square-error (MSE) between original dense weights and sparse weights, rather than to keep the total magnitude of weights unchanged.

\paragraph{Freezing scaling factor}
As \citet{vanderschueren2022straightthrough} use a dynamic $\beta$ for every iteration, we argue that this doesn't align with our approach. We explain our solutions in two parts.

\begin{figure}[h]
    \centering
    \includegraphics[width=1\linewidth]{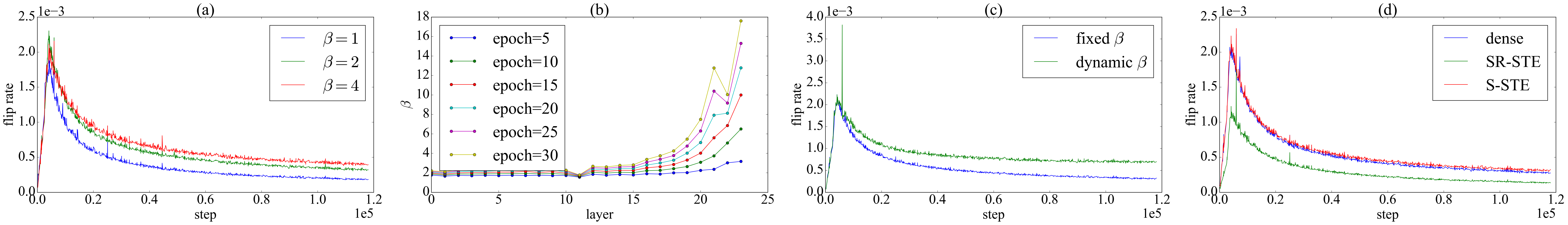}
    \caption{(a) Flip rate curve over the training process with different $\beta$ on Transformer-base. (b) Dynamically recalculated $\beta$ at each layer on different epochs. Results show that frequently updating $\beta$ will cause it to be unexpectedly large. (c) Flip rate curve over the training process with fixed and dynamic $\beta$ on Transformer-base. (d) Flip rate of dense, SR-STE and S-STE algorithm on Transformer-base.}
    \label{fig:3456}
\end{figure}

First, we find it interesting that $\beta$ have a subtle correlation with flip rate: \textit{in a sparse model, larger $\beta$ usually results in higher flip rate.} The reason can be explained by the accuracy of gradients. As we use STE in the backward pass, the approximation $\nabla_{\wv} f\approx \nabla_{\tilde \wv} f$ is valid when $\tilde \wv$ and $\wv$ are close enough. However, if scale is too large then optimal, $\wv$ and $\tilde \wv$ are too far apart to guarantee this. Such a mismatch leads to incorrectness in descending directions, and thus unstability in optimization increase flip rate; see Fig. \ref{fig:3456}(a).

Second, we argue that dynamically computing scaling factor for each iteration leads to high flip rate in our training process. Fig. \ref{fig:3456}(b) shows the results dynamically changing $\beta$ will make it increase with iterations, especially for later layers. Fig. \ref{fig:3456}(c) shows flip rate of this network, which has a significantly higher tail than the dense one. Considering high flip rate is harmful, we propose to compute scaling factor $\beta$ only in the first iteration. After that, we use the same $\beta$ in the rest of the training. Fig. \ref{fig:3456}(d) shows the flip rate of our fixed scaling S-STE, which perfectly aligns with the dense one.

\paragraph{Minimizing MSE} \citet{vanderschueren2022straightthrough} choose to scale up $S_{soft}(\wv)$ to have the same L1-norm as $\wv$: $\beta = \norm{\wv}_1 / \norm{S_{soft}(\wv)}_1$. However, we choose to minimize the MSE gap of $S_{soft}(\wv)$ and $\wv$. As \cite{chmiel2023minimum} point out, sparsifying weights in the forward pass should minimize MSE rather than an unbiased estimation. In our method, to determine an optimal scale $\beta$, we need to minimize
\begin{align}
    \text{MSE} = \norm{\wv-\beta S_{soft}(\wv)}^2=\norm{\wv}^2-2\wv^\top S_{soft}(\wv) \beta+ \norm{ S_{soft}(\wv)}^2 \beta^2.
\end{align}
Rearrange the terms and taking partial derivative of $\beta$, we choose $\beta = \wv^\top S_{soft}(\wv) / \norm{ S_{soft}(\wv)}^2$.
The comparison between no scaling, our minimizing MSE and keeping L1-norm can be found in Table \ref{table:4}. Result show that our method yields the best results in practice.

\begin{table}[ht]
\caption{Experimental result of different $\beta$ on Transformer-base.}
\label{table:4}
\begin{center}
\begin{tabular}{llll}
\toprule
$\beta$ Recipe & Test BLEU & Val loss & Avg epoch loss \\
              \midrule
No scaling & 25.28 &  4.044 &   4.670 \\
Keeping L1-norm same \cite{vanderschueren2022straightthrough} & 25.85       &  4.019 &   4.627 \\
\textbf{Minimizing MSE (S-STE)} & \bm{$26.3$} &  \bm{$4.007$} & \bm{$4.605$} \\
\bottomrule
\end{tabular}
\end{center}
\end{table}




%% file: 5-implementation.tex
\section{Other implementation skills}
\subsection{Minimum-variance unbiased estimation}
\label{sec:5.1}

To accelerate the backward propagation, \citet{chmiel2023minimum} suggest using a minimum-variance unbiased estimator (MVUE). For every linear layer $\Zv_l = \Xv_{l} S(\Wv_l)^\top$, there are two matrix multiplications of the backward pass in total: $\nabla_{\Xv_l} = \nabla_{\Zv_l} S(\Wv_l)$ and $\nabla_{\Wv_l} = \nabla_{\Zv_l}^{\top} \Xv_l$, where $\Xv_l$ is the input of the $l$-th layer, $\Wv_l$ and $\Zv_l$ are the weight matrix and output activation. We conduct MVUE on both two matrix multiplications and compare their results: $\nabla_{\Xv_l} = \nabla_{\Zv_l} \operatorname{MVUE}(S(\Wv_l)^\top)^\top$ and $\nabla_{\Wv_l} = \operatorname{MVUE}(\nabla_{\Zv_l}^{\top}) \Xv_l$. Specifically, we choose $S(\Wv_l)$ and $ \nabla_{\Zv_l} $ because they both have built-in sparsity \cite{lin2020deep}.
However, we only choose to sparsify the latter one. Firstly, it is proven by \citet{hu2024accelerating} and \citet{chmiel2023minimum} that minimum loss of accuracy is guaranteed for MVUE on $\nabla_{\Zv_l}$. Secondly,
using MVUE on $S(\Wv_l)$ will make errors accumulate along the back propagation, and results in large standard deviation of gradient for the first few layers. Besides, results in Table \ref{table:6} also show minimum loss of accuracy of MVUE on $ \nabla_{\Zv_l} $ while obvious accuracy loss on $S(\Wv_l)$. Thus, we choose to sparsify only $\nabla_{\Zv_l}$ in the backward pass.

\begin{table}
\caption{Results of different MVUE strategies on GPT-2 774M with 4000 steps. Sparsifying $S(\Wv)^\top$ introduces huge loss of accuracy while sparsifying $\nabla_{\Zv}^\top$ is acceptable with little loss.}
\label{table:6}
\begin{center}
\begin{tabular}{lllll}
\toprule
S-STE & $\operatorname{MVUE}(S(\mathbf{W})^\top)$ & $\operatorname{MVUE}(\nabla_{\mathbf{Z}}^\top)$ & comment & loss      \\
\midrule
- & \XSolidBrush & \XSolidBrush & dense   & 3.3948    \\
- & \XSolidBrush & \XSolidBrush & SR-STE  & 3.4739    \\
\Checkmark & \XSolidBrush & \XSolidBrush & & 3.4333    \\
\Checkmark & \Checkmark & \XSolidBrush & & 3.4644    \\
\Checkmark & \Checkmark & \Checkmark & & 3.4773    \\
\Checkmark & \XSolidBrush & \Checkmark & & \textbf{3.4480}  \\
\bottomrule
\end{tabular}
\end{center}
\end{table}

\subsection{FP8 training}
To further accelerate pre-training of networks, we utilize popular FP8 workflow in training. Similar to Transformer Engine \footnote{\url{https://github.com/NVIDIA/TransformerEngine}}, we use FP8 e3m4 in forward pass and e5m2 in backward pass. Besides, we use per-tensor rescaling before casting to FP8 formats.

\paragraph{Theoretical acceleration of S-STE}
While 2:4 sparsity can accelerate GEMMs up to 2x faster, FP8 quantization can accelerate an additional 2x on this basis. Thus, the three GEMMs in Sec. \ref{sec:5.1} can be 4x, 2x, 4x faster. To sum up, we have theoretically 3x faster in forward and backward pass.

%% file: 6-experiments.tex
\section{Experiments}

We validate the feasibility of our proposed method S-STE on machine translation (Transformer \cite{vaswani2023attention}), image classification (DeiT \cite{touvron2021training}) and generative large language models (GPT-2 \cite{Radford2019LanguageMA} series). For all models, we replace the two linear layers in the feed forward network of each transformer block with S-STE. We keep the rest of the networks, the optimization algorithms as well as all hyperparameters the same as their dense counterparts.

For Transformer, we train Transformer-base models on WMT 14 En-De dataset  \cite{Bojar2014FindingsOT} with fairseq \cite{ott2019fairseq} codebase and evaluate it with BLEU \cite{article} scores. For DeiT, we pre-train Deit-small model for ImageNet-1K \cite{5206848} classification task. For GPT-2, we pre-train GPT-2 124M, 350M and 774M models on OpenWebText \cite{Gokaslan2019OpenWeb} and evaluate it on GLUE \cite{wang2019glue} and SQuAD \cite{rajpurkar2016squad} benchmarks. We also compare our method with state-of-the-art 2:4 training methods (SR-STE \cite{zhou2021learning}, Bi-Mask \cite{zhang2023bidirectional} and STEP \cite{lu2023step}). The pre-training and evaluation scripts are publicly available at \url{https://github.com/thu-ml/2by4-pretrain-acc-examples}.

\paragraph{Machine translation}
We first apply S-STE to train a 12-layer Transformer-base and compare it with SR-STE and STEP. Note that we use fairseq codebase with SacreBleu metric, whose baseline should be 26.5 (the result of our reproduction is 26.42). The results are shown in Table \ref{table:wmt}. Compared with SR-STE, our method improves by 0.3 and 0.5 on test set and validation set respectively, which is the closest to baseline. Besides, we improve by 0.6 compared to STEP on test set.

\begin{table}[!ht]
\centering
\caption{Experimental Results for Transformer-base on En-De dataset.}
\label{table:wmt}
\begin{center}
\begin{tabular}{lllll}
\toprule
Method & Avg epoch loss& Test BLEU & Val BLEU & Val loss \\
\midrule
Dense & 4.555 & 26.42 & 26.49 & 3.977 \\
SR-STE &  4.61 & 25.84 & 26.08 & 4.023 \\
STEP &  4.682 & 25.52 & 26.01 & 4.085 \\
\textbf{S-STE} & \bm{$4.617$} & \bm{$26.11$} &  \bm{$26.53$} & \bm{$4.011$} \\
\bottomrule
\end{tabular}
\end{center}
\end{table}

\begin{table}[!ht]
\centering
\caption{Experimental Results for DeiT-small on ImageNet-1k. The Bi-Mask and SR-STE results are from \cite{zhang2023bidirectional}.}
\label{table:deit}
\begin{center}
\begin{tabular}{llll}
\toprule
Model & Method & Test acc1 & Test acc5  \\
\midrule
\multirow{3}{*}{DeiT-tiny} & Dense & 72.2 & 91.1  \\
&SR-STE \cite{zhang2023bidirectional} & 67.8 & 88.6 \\
&\textbf{S-STE} & \bm{$68.5$} & \bm{$88.9$}  \\
\midrule
\multirow{4}{*}{DeiT-small} & Dense & 79.9 & 95  \\
& SR-STE \cite{zhang2023bidirectional} & 75.7 & - \\
& Bi-Mask \cite{zhang2023bidirectional} & 77.6 & -  \\
& \textbf{S-STE} & \bm{$78.5$} & \bm{$94.4$}  \\
\bottomrule
\end{tabular}
\end{center}
\end{table}

\paragraph{Image classification}
We further investigate the effectiveness of S-STE to train DeiT-tiny and DeiT-small on ImageNet-1k; see Table \ref{table:deit}. Results show S-STE also achieve the best performance among different methods, with only has 1.4\% degradation from the dense model. Notably, S-STE surpasses SOTA 2:4 training method Bi-Mask on this task (0.9\% top1 accuracy improvement) and popular SR-STE method (2.8\% top1 accuracy improvement).

\paragraph{Generative language models}
We compare S-STE with dense, normal SR-STE and SR-STE with dense fine-tuning \cite{hu2024accelerating} (SR-STE+DF) models on GLUE and SQuAD tasks. 
The SR-STE+DF models first use SR-STE to train a 2:4 sparse model, and switch to dense training for the last 1/6 iters of pre-training (which stands for ``dense fine-tune''). In downstream tasks it also use dense parameters to make predictions, similar to dense models. Results in Table \ref{table:downstream} and \ref{table:fullglue} show that S-STE completely surpasses SR-STE on both tasks. Even for SR-STE+DF models, S-STE still have an advantage, with an improvement of 1.5 on GLUE average and 1.2/0.9 on SQuAD for GPT-2 774M.

\begin{table}[!ht]
\caption{SQuAD and GLUE scores of different sizes and pre-training methods on GPT-2. We use 2:4 sparse weights to evaluate S-STE model, while dense parameters to evaluate the rest. Of note, SR-STE denotes the original SR-STE workflow (without backward MVUE), and ``T-SR-STE+DF'' denotes the combination of transposable SR-STE \& backward MVUE \& sparse-dense training workflow, proposed by \citet{hu2024accelerating}. S-STE settings here include backward MVUE \& FP8 training.}
\label{table:downstream}
\begin{center}
\begin{tabular}{lllllll}
\toprule

\multirow{2}{*}{Params} & \multirow{2}{*}{Pre-train} & \multirow{2}{*}{Fine-tune} & \multirow{2}{*}{\makecell{Pre-train\\val loss}} & \multicolumn{2}{l}{SQuAD} & \multirow{2}{*}{GLUE@Avg} \\
\cmidrule(lr){5-6} & & & & EM & F1 \\
\midrule
\multirow{5}{*}{124M} & Dense  & Dense & 2.907 & 67.6 & 78.8 & $73.9 \pm 1.1$ \\
& T-SR-STE+DF \cite{hu2024accelerating}  & Dense & 2.952 & 67.5 & 78.5 & {$74.3 \pm 0.5$} \\
& T-SR-STE & Dense & 3.076 & 66.3 & 77.2 & $72.6 \pm 0.2$ \\
& SR-STE & Dense & 2.982 & 66.2 & 77.5 & $73.8 \pm 0.3$ \\
& \textbf{S-STE} & \textbf{S-STE} & \textbf{2.984} & \textbf{68} & \textbf{78.8} & \bm{$74.1 \pm 0.4$} \\
\midrule
\multirow{5}{*}{350M} & Dense & Dense & 2.618 & 73.2 & 83.6 & $76.3 \pm 0.1$ \\
& T-SR-STE+DF \cite{hu2024accelerating} & Dense & 2.688 & 71.9 & 82.4 & $77.1 \pm 0.2$ \\
& T-SR-STE & Dense & 2.718 & 72.3 & 82.6 & $76.3 \pm 0.4$ \\
& SR-STE & Dense & 2.690 & 72.0 & 82.4 & $76.8 \pm 0.4$\\
& \textbf{S-STE} & \textbf{S-STE} &  \textbf{2.713} & \textbf{72.2} & \textbf{82.7} & \bm{$76.9 \pm 0.6$}\\
\midrule
\multirow{3}{*}{774M} & Dense & Dense  & 2.493 & 74.3  & 84.9 & $76.2 \pm 0.4$ \\
& T-SR-STE+DF \cite{hu2024accelerating} & Dense  & 2.564 & 74.3 & 84.6 & $77.1 \pm 0.4$ \\
& \textbf{S-STE} & \textbf{S-STE} & \textbf{2.547} & \textbf{75.5} & \textbf{85.5} & \bm{$78.6 \pm 0.8$}\\
\bottomrule
\end{tabular}
\end{center}
\end{table}

\begin{table}[!ht]
\centering
\caption{Different fine-tuning results on GLUE and SQuAD.}
\label{table:finetuning}
\begin{tabular}{lllll}
\toprule
Model & Downstream task & Pre-train & Fine-tune & Avg score \\
\midrule
\multirow{4}{*}{GPT-2 124M} & GLUE & S-STE & Hard-thresholding & $73.9\pm0.6$ \\
& \textbf{GLUE} & \textbf{S-STE} & \textbf{S-STE} & \bm{$74.1\pm0.4$} \\
& SQuAD & S-STE & Hard-thresholding & $67.6/78.6$ \\
& \textbf{SQuAD} & \textbf{S-STE} & \textbf{S-STE} & \bm{$68/78.8$} \\
\bottomrule
\end{tabular}
\end{table}

\paragraph{Fine-tuning} We illustrate the viability of S-STE for fine-tuning a pre-trained model, presenting a coherent workflow of accelerating both training and inference (dense fine-tuning cannot produce a sparse model for inference acceleration); see Table \ref{table:finetuning}, \ref{table:downstream}.

\paragraph{Ablation study}
In this part, We explore the effectiveness of S-STE, MVUE, and FP8 separately. We pre-train DeiT-small model on ImageNet-1K dataset for image classification. Combinations of these partitions in Table \ref{table:ablation} show that: 1) FP8 training has little affect on pre-training accuracy (0.2\% of acc1); 2) MVUE leads to minimal loss of performance (0.1\% of acc1).

\begin{table}[ht]
\caption{Experimental result of S-STE (soft-thresholding and weight rescaling), MVUE and FP8 training with DeiT-small on ImageNet-1K.}
\label{table:ablation}
\begin{center}
\begin{tabular}{lllllll}
\toprule
\makecell{Soft-\\thresholding} & \makecell{Weight\\rescaling} & $\operatorname{MVUE}(\nabla_{\mathbf{Z}}^\top)$ & FP8 & Comment & Test acc1 & Test acc5 \\
\midrule
- & - & \XSolidBrush & \XSolidBrush & Dense & 79.9 & 95 \\
- & - & \XSolidBrush & \Checkmark & Dense; FP8 & 79.7 & 94.9 \\
\XSolidBrush & \XSolidBrush & \XSolidBrush & \XSolidBrush & Hard-thresholding & 77.7 & 93.9 \\
\Checkmark & \Checkmark & \XSolidBrush & \XSolidBrush & & 78.8 & 94.6 \\
\Checkmark & \XSolidBrush & \XSolidBrush & \XSolidBrush & & 78.9 & 94.7 \\
\Checkmark & \Checkmark & \XSolidBrush & \Checkmark & & 78.6 & 94.4 \\
\Checkmark & \Checkmark & \Checkmark & \XSolidBrush & & 78.9 & 94.6 \\
\Checkmark & \XSolidBrush & \Checkmark & \XSolidBrush & & 78.2 & 94.2 \\
\Checkmark & \Checkmark & \Checkmark & \Checkmark & & \textbf{78.5} & \textbf{94.4} \\
\bottomrule
\end{tabular}
\end{center}
\end{table}

\paragraph{Acceleration}
For acceleration, we measure the acceleration ratio of a typical GPT-2 model using
implementation from \citet{hu2024accelerating}. Note that on H100 GPUs, FP8 2:4-spMM kernel turns out to be unsatisfying; see Appendix \ref{limitations}. Consequently, we fall back to use RTX3090 GPUs with FP16 training. For inference, we achieve 1.53x speedup with FFN layer and 1.23x speedup with the network; for pre-training, we achieve 1.32x speedup for FFN layer and 1.18x speedup for the network (Appendix \ref{sec:acceleration}).

%% file: 7-related.tex
\section{Related work}
\paragraph{Unstructured pruning and coarse-grained structured pruning}
Pruning is to remove redundant weights from the dense model. Traditional one-shot pruning methods \cite{han2015learning,han2016deep,frankle2018the,frankle2020stabilizing,mishra2021accelerating,lee2019snip} and dynamic sparse training methods \cite{evci2021rigging,chen2020lottery,chen2021earlyBERT, you2022drawing} mostly target on unstructured sparsity. While most of them have acceleration effect on CPUs, it is hard for these methods to work well on modern GPUs. Coarse-grained structured sparsity \cite{yin2023dynamic,lasby2023dynamic,he2017channel,lagunas2021block} takes effect to acceleration, but since they often remove a whole chennel or a block, loss of information is non-negligible.

\paragraph{Fine-grained N:M sparsity for inference and pre-training}
Among all pruning techniques for pre-training, N:M sparsity is a promising approach towards accelerating large models, which is also known as fine-grained structured sparsity. Nvidia demonstrates 2x theoretical speedup on its Ampere GPUs with 2:4 sparsity for post-training \cite{mishra2021accelerating} and inference \cite{sun2023wanda,frantar2023sparsegpt,Pandey2007RIAAR,dathathri2020plug}. To leverage this property to accelerate pre-training, a number of approaches and their improvements are proposed \cite{zhou2021learning,lu2023step,zhang2023bidirectional,bambhaniya2024progressive,chmiel2023minimum,hubara2021accelerated,hu2024accelerating}. However, all these methods are based on a discontinuous pruning function that is hard to optimize and results in unsatisfactory accuracy, which we will discuss in this study.

\paragraph{FP8 quantization} While 16-bit float tensors are widely used in pre-training, FP8 – where float numbers stored in 8 bits – is a popular quantization methods which theoretically accelerates GEMMs up to 4x faster than its fp32 counterparts and 2x faster than its FP16/BF16 counterparts \cite{micikevicius2022fp8,perez2023training,vanbaalen2023fp8,wang2018training,noune20228bit}. With e3m4 data format used in forward and e5m2 format \cite{Sun2019Hybrid8F} in backward, pre-trained models can achieve minimum loss of accuracy while greatly boosting the efficiency of training.

%% file: 8-conclusion.tex
\section{Conclusions and future work}
In this study we discuss the importance of pruning continuity in effective 2:4 sparse pre-training. We analyse the drawback of traditional hard-thresholding pruning function and its variation (SR-STE) and argue that the main limits being discontinuity. Based on our analysis and soft-thresholding for channel pruning, we propose S-STE, which prunes weights in a continuous manner. Experiments show that our method surpasses previous state-of-the-art methods on a wide range of tasks.

Our proposed S-STE approach primarily targets linear layers within FFN networks. Nevertheless, QKV projection layers necessitate further exploration to devise an effective dynamic sparse training strategy that harmonizes with attention mechanisms. Furthermore, our current choice of continuous pruning function represents only one possible solution; alternative, smoother pruning functions may be necessary to achieve improved continuity and mitigate potential discontinuities.

%% file: appendix.tex
\section{Appendix / supplemental material}

\subsection{Proof of Theorem \ref{thm:1}}
\begin{proof}
\label{proof:1}

We prove this by demonstrating $S_{soft}$ is continuous on every 4-element block. Of note, suppose $\av = [a_1,a_2,a_3,a_4]^\top$. Assume, without loss of generality, that $|a_1|\leq |a_2| \leq |a_3| \leq |a_4|$. Our goal is to prove $\forall \epsilon>0, \exists \delta>0, \st \text{ when } |a_1'-a_1|<\delta, |a_2'-a_2|<\delta, |a_3'-a_3|<\delta \text{ and } |a_4'-a_4|<\delta, |(S_{soft}(\av'))_i-a_i|<\epsilon$, where $\av' = [a_1',a_2',a_3',a_4']^\top$.

1) We start from the simplest case where $|a_1| < |a_2| < |a_3| < |a_4|$. Then we have
\begin{equation*}
    S_{soft} = [0,0,\operatorname{sign}(a_3)(|a_3|-|a_2|),\operatorname{sign}(a_4)(|a_4|-|a_2|)].
\end{equation*}
This order holds when
\begin{equation*}
    \delta < \frac{1}{2}\min \{(|a_2|-|a_1|),|a_3|-|a_2|),|a_4|-|a_3|)\}.
\end{equation*}
Thus,
\begin{equation*}
    S_{soft}(\av') = [0,0,\operatorname{sign}(a_3')(|a_3'|-|a_2'|),\operatorname{sign}(a_4')(|a_4'|-|a_2'|)].
\end{equation*}
The signs of $\av$ is unchanged when
\begin{equation*}
    \delta < \min \{|a_1|,|a_2|,|a_3|,|a_4|\}.
\end{equation*}
We have
\begin{align*}
    &\operatorname{sign}(a_4')(|a_4'|-|a_2'|)-\operatorname{sign}(a_4)(|a_4|-|a_2|) \\
    &\leq |~|a_4'|-|a_2'| -|a_4|+|a_2|~| \\
    &\leq |~|a_4'|-|a_4|~| + |~|a_2'|-|a_2|~| \\
    &\leq 2\delta
\end{align*}
Take $\delta \leq \frac{1}{2}\epsilon$ and this is done. It is similar to prove that $\operatorname{sign}(a_3')(|a_3'|-|a_2'|)-\operatorname{sign}(a_3)(|a_3|-|a_2|)\leq \epsilon$ using the same method.

2) We then consider the cases where there are two equivalents in $\av$. If $|a_1| = |a_2| < |a_3| < |a_4|$ or $|a_1| < |a_2| < |a_3| = |a_4|$, the proof should follow 1) as no flip happens. Thus we only consider the situation where $|a_1| < |a_2| = |a_3| < |a_4|$. Under these circumstances, a flip will happen on the second and third dimensions of $\av$.
\begin{align*}
     S_{soft}(\av) = [0,0,0,\operatorname{sign}(a_4)(|a_4|-|a_2|)]
\end{align*}
Without loss of generality we assume $|a^\prime_1| < |a'_2| \leq |a'_3| < |a'_4|$. Thus,
\begin{align*}
    S_{soft}(\av') = [0,0,\operatorname{sign}(a_3')(|a_3'|-|a_2'|),\operatorname{sign}(a_4')(|a_4'|-|a_2'|)].
\end{align*}
The proof of the fourth dimension is similar to 1), so we only focus on $a_3$.
\begin{align*}
    &\operatorname{sign}(a_3')(|a_3'|-|a_2'|) \\
    &\leq |~|a_3'|-|a_2'|~| \\
    &\leq |~|a_3'|-|a_3|~| + |~|a_2'|+|a_2|~| \\
    &\leq 2\delta
\end{align*}
Take $\delta \leq \frac{1}{2}\epsilon$ and this is done.

3) If there exists three or four equivalent in $\av$, a flip will happen at the second and third dimension. Thus, these cases can be reduced to 1) or 2).

\end{proof}

\subsection{GLUE scores of GPT-2}

See Table \ref{table:fullglue}.

\begin{table}[!ht]
\caption{Comparison between GLUE scores of different pre-train methods on GPT-2 models. This table is the elaboration of Table \ref{table:downstream}.}
\label{table:fullglue}
\begin{center}
\resizebox{\textwidth}{!}{
\begin{tabular}{lllllllllllll}
\toprule
Params & Method & Avg score & CoLA & MNLI & MRPC & QNLI & QQP & RTE & SST-2 & STS-B & WNLI \\
\midrule
\multirow{5}{*}{124M} & Dense    & $73.9 \pm 1.1$ & $44.6 \pm 0.9$ & $82.0 \pm 0.1$ & $78.3 \pm 1.3/84.8 \pm 1.0$ & $88.4 \pm 0.2$ & $90.0 \pm 0.0$ & $86.5 \pm 0.0/61.3 \pm 1.5$& $91.9 \pm 0.2$ & $77.3 \pm 3.2/77.9 \pm 2.9$ & $24.3 \pm 7.1$\\

& {T-SR-STE+DF \cite{hu2024accelerating}}&{$74.3 \pm 0.5$}&{$44.8 \pm 1.3$}&{$81.5 \pm 0.2$}&{$77.5 \pm 1.8/84.2 \pm 1.3$}&{$87.8 \pm 0.1$}&{$89.5 \pm 0.1$}&{$85.9 \pm 0.1/66.0 \pm 1.0$}&{$90.6 \pm 0.4$}&{$80.0 \pm 0.8/80.3 \pm 0.5$}&{$23.9 \pm 6.4$}\\

& {T-SR-STE}&{$72.6 \pm 0.2$}&{$41.9 \pm 0.3$}&{$81.0 \pm 0.2$}&{$76.3 \pm 0.9/83.4 \pm 0.7$}&{$87.0 \pm 0.3$}&{$89.3 \pm 0.1$}&{$85.6 \pm 0.1/60.6 \pm 3.4$}&{$90.9 \pm 0.4$}&{$76.2 \pm 3.2/76.5 \pm 3.0$}&{$21.8 \pm 4.4$}
\\

& {SR-STE}&{$73.8 \pm 0.3$}&{$38.3 \pm 2.8$}&{$80.9 \pm 0.2$}&{$79.7 \pm 0.7/85.9 \pm 0.6$}&{$87.1 \pm 0.3$}&{$89.5 \pm 0.1$}&{$85.8 \pm 0.2/65.5 \pm 1.0$}&{$90.5 \pm 0.4$}&{$80.9 \pm 1.2/80.9 \pm 1.1$}&{$20.1 \pm 1.8$}
\\

& \textbf{S-STE} & \bm{ $74.1 \pm 0.4$ }&\bm{ $42.3 \pm 1.1$ }&\bm{ $80.5 \pm 2.8$ }&\bm{ $79.3 \pm 1.9/85.6 \pm 1.4$ }&\bm{ $88.1 \pm 0.2 $}&\bm{ $89.8 \pm 0.1$ }&\bm{ $86.2 \pm 0.1/62.9 \pm 1.1$ }&\bm{ $91.9 \pm 0.4$ }&\bm{ $81.0 \pm 1.1/81.2 \pm 1.0$ }&\bm{ $20.8 \pm 3.6$} \\

\midrule
\multirow{5}{*}{350M} & Dense& $76.3 \pm 0.1$ & $54.3 \pm 0.4$ & $85.1 \pm 0.1$ & $80.7 \pm 1.0/86.6 \pm 0.7$ & $90.7 \pm 0.1$ & $91.0 \pm 0.1$ & $87.8 \pm 0.1/64.9 \pm 1.7$ & $93.5 \pm 0.4$ & $81.7 \pm 1.2/82.2 \pm 0.8$ & $17.6 \pm 3.2$\\

&{T-SR-STE+DF \cite{hu2024accelerating}}&{$77.1 \pm 0.2$}&{$51.8\pm1.8$}&{$84.3\pm0.1$}&{$80.6\pm1.3/86.5\pm0.8$}&{$90.4\pm0.2$}&{$90.7\pm0.1$}&{$87.5\pm0.1/66.7\pm1.3$}&{$93.3\pm 0.4$}&{ $83.4\pm1.1/83.5\pm1.1$}&{$26.4 \pm 4.0$}\\

&{T-SR-STE}&{$76.3 \pm 0.4$}&{$50.0 \pm 1.7$}&{$84.1 \pm 0.2$}&{$81.4 \pm 1.5/87.1 \pm 1.1$}&{$90.0 \pm 0.3$}&{$90.6 \pm 0.1$}&{$87.3 \pm 0.1/67.9 \pm 1.5$}&{$93.3 \pm 0.4$}&{$81.3 \pm 1.5/81.4 \pm 1.4$}&{$20.6 \pm 3.8$}\\
      
&{SR-STE}&{$76.8 \pm 0.4$}&{$47.2 \pm 3.0$}&{$84.3 \pm 0.2$}&{$81.4 \pm 0.9/87.2 \pm 0.6$}&{$90.2 \pm 0.1$}&{$90.8 \pm 0.1$}&{$87.6 \pm 0.1/68.3 \pm 1.4$}&{$93.9 \pm 0.1$}&{$82.0 \pm 1.6/82.0 \pm 1.7$}&{$27.1 \pm 3.1$}\\

&\textbf{S-STE}&\bm{$76.9 \pm 0.6$ }&\bm{ $54.2 \pm 1.7$ }&\bm{ $84.6 \pm 0.2$ }&\bm{ $80.2 \pm 1.3/86.1 \pm 0.9$ }&\bm{ $90.5 \pm 0.3$ }&\bm{ $90.8 \pm 0.1$ }&\bm{ $87.5 \pm 0.2/65.1 \pm 1.9$ }&\bm{ $93.7 \pm 0.4$ }&\bm{ $83.6 \pm 1.1/83.8 \pm 1.1$ }&\bm{$22.5 \pm 3.9$}\\

\midrule
\multirow{3}{*}{774M} & Dense & $76.2 \pm 0.4$ & $57.5 \pm 2.0$ & $86.1 \pm 0.1$ & $80.3 \pm 1.3/86.4 \pm 0.9$ & $91.4 \pm 0.2$ & $91.1 \pm 0.1$ & $88.0 \pm 0.1/67.7 \pm 2.6$ & $94.6 \pm 0.4$ & $77.3 \pm 3.3/78.4 \pm 2.9$ & $15.1 \pm 2.3$\\

&{T-SR-STE+DF \cite{hu2024accelerating}}&{$77.1 \pm 0.4$}&{$55.9 \pm 0.9$}&{$85.6 \pm 0.2$}&{$81.2 \pm 0.6/87.0 \pm 0.4$}&{$91.4 \pm 0.1$}&{$91.0 \pm 0.1$}&{$87.8 \pm 0.1/71.5 \pm 0.7$}&{$94.2 \pm 0.4$}&{$81.8 \pm 1.3/82.3 \pm 1.2$}&{$15.8 \pm 1.2$}\\

&\textbf{S-STE}&\bm{$78.6 \pm 0.8$ }&\bm{ $57.3 \pm 2.7$ }&\bm{ $86.6 \pm 0.2$ }&\bm{ $80.6 \pm 1.4/86.6 \pm 0.9$ }&\bm{ $92.0 \pm 0.1$ }&\bm{ $91.5 \pm 0.1$ }&\bm{ $88.5 \pm 0.1/78.3 \pm 1.5$ }&\bm{ $94.9 \pm 0.3$ }&\bm{ $85.5 \pm 1.2/85.7 \pm 1.1$ }&\bm{ $16.1 \pm 5.9$}\\

\bottomrule
\end{tabular}
}
\end{center}
\end{table}

\subsection{Acceleration}
\label{sec:acceleration}
See Table \ref{table:pretrainAcc}.

\begin{table}[!ht]
\centering
\caption{Pre-training acceleration ratio with different different batch size $N$, sequence length $n$, embedding dimension $d$ and heads number $h$ on single FFN block and transformer block of GPT-2 with RTX 3090 GPUs.}
\label{table:pretrainAcc}
\begin{tabular}{lllllll}
\toprule
& N  & n    & d    & h  & FFN & GPT-2 \\
\midrule
\multirow{5}{*}{Pre-train} & 4  & 2048 & 5120 & 40 & 1.31      & 1.18        \\
& 16 & 2048 & 7168 & 56 & 1.32      & 1.18         \\
& 8  & 2048 & 7168 & 56 & 1.33      & 1.17        \\
& 4  & 2048 & 7168 & 56 & 1.31      & 1.17        \\
& 4  & 2048 & 9216 & 72 & 1.31      & 1.18       \\
\midrule
\multirow{2}{*}{Inference} & 16 & 2048 & 7168 & 56 & 1.54 & 1.23 \\
& 8  & 2048 & 7168 & 56 & 1.46 & 1.15 \\
\bottomrule
\end{tabular}
\end{table}

\subsection{Limitations}
\label{limitations}
As we propose accuracy results of S-STE on several tasks, no actual acceleration result is given. While theoretically 2x faster results can be expected (FP8 quantization), the NVIDIA acceleration library (cuSPARSElt \cite{mishra2021accelerating}) is not satisfactory, which causes inconvenience on implementation. The peak FLOPS of 2:4-spMM is lower than theoretical GEMM FLOPS; see Table \ref{table:h100}.

\begin{table}[!h]
\caption{Peak FLOPS of general matrix multiplications (GEMMs) and 2:4 sparse matrix multiplications (2:4-spMMs) on H100. The size we take to test is $16384\times16384\times16384$.}
\label{table:h100}

\begin{center}
\begin{tabular}{lll}
\toprule
& GPU & FP8 Tensor Core \\
\midrule
\multirow{4}{*}{Specifications} & H100 PCIe 2:4-spMM & 3200 TFLOPS \\
 & H100 PCIe GEMM     & 1600 TFLOPS \\
 & H100 SXM 2:4-spMM  & 4000 TFLOPS \\
 & H100 SXM GEMM      & 2000 TFLOPS \\
\midrule
\multirow{2}{*}{\makecell{Actual results \\ with cuSPARSElt}} & H100 SXM 2:4-spMM  & 1900 TFLOPS     \\
                                                & H100 SXM GEMM      & 1500 TFLOPS  \\ \bottomrule 
\end{tabular}
\end{center}
\end{table}

\begin{table}[!h]
\caption{GPU Hours of pre-training models on RTX 4090.}
\label{table:resource}
\begin{center}
\begin{tabular}{ll}
\toprule
                 & GPU Hours \\
                 \midrule
GPT-2 124M       & 400       \\
GPT-2 350M       & 900       \\
GPT-2 774M       & 2500      \\
Transformer-base & 30        \\
DeiT-base        & 120      \\
\bottomrule
\end{tabular}
\end{center}
\vskip -0.1in
\end{table}

\subsection{Broader Impact}
S-STE can be used mainly to accelerate the pre-training stage of large-scale networks, like LLaMA \cite{touvron2023llama} and GPT-4 \cite{openai2024gpt4}. Theoretically GEMMs of the FFN layer can be accelerated up to 4x faster than FP16 dense models, which would greatly reduce the electric power consumption of pre-training modern large-scale models. However, this method may also used for some models that is non-compliance with regulations and ethics, such as models that generate discriminatory contents.

\subsection{Experiments compute resources}
To replicate our experiments, we provide the estimated GPU hours of each setting; see Table \ref{table:resource}.


%% file: main.bbl
\begin{thebibliography}{53}
\providecommand{\natexlab}[1]{#1}
\providecommand{\url}[1]{\texttt{#1}}
\expandafter\ifx\csname urlstyle\endcsname\relax
  \providecommand{\doi}[1]{doi: #1}\else
  \providecommand{\doi}{doi: \begingroup \urlstyle{rm}\Url}\fi

\bibitem[Bambhaniya et~al.(2024)Bambhaniya, Yazdanbakhsh, Subramanian, Kao, Agrawal, Evci, and Krishna]{bambhaniya2024progressive}
Abhimanyu~Rajeshkumar Bambhaniya, Amir Yazdanbakhsh, Suvinay Subramanian, Sheng-Chun Kao, Shivani Agrawal, Utku Evci, and Tushar Krishna.
\newblock Progressive gradient flow for robust n:m sparsity training in transformers, 2024.

\bibitem[Bengio et~al.(2013)Bengio, Léonard, and Courville]{bengio2013estimating}
Yoshua Bengio, Nicholas Léonard, and Aaron Courville.
\newblock Estimating or propagating gradients through stochastic neurons for conditional computation, 2013.

\bibitem[Bojar et~al.(2014)Bojar, Buck, Federmann, Haddow, Koehn, Leveling, Monz, Pecina, Post, Saint-Amand, Soricut, Specia, and Tamchyna]{Bojar2014FindingsOT}
Ondrej Bojar, Christian Buck, Christian Federmann, Barry Haddow, Philipp Koehn, Johannes Leveling, Christof Monz, Pavel Pecina, Matt Post, Herve Saint-Amand, Radu Soricut, Lucia Specia, and Ales Tamchyna.
\newblock Findings of the 2014 workshop on statistical machine translation.
\newblock In \emph{WMT@ACL}, 2014.
\newblock URL \url{https://api.semanticscholar.org/CorpusID:15535376}.

\bibitem[Bottou et~al.(2018)Bottou, Curtis, and Nocedal]{bottou2018optimization}
Léon Bottou, Frank~E. Curtis, and Jorge Nocedal.
\newblock Optimization methods for large-scale machine learning, 2018.

\bibitem[Brown et~al.(2020)Brown, Mann, Ryder, Subbiah, Kaplan, Dhariwal, Neelakantan, Shyam, Sastry, Askell, Agarwal, Herbert-Voss, Krueger, Henighan, Child, Ramesh, Ziegler, Wu, Winter, Hesse, Chen, Sigler, Litwin, Gray, Chess, Clark, Berner, McCandlish, Radford, Sutskever, and Amodei]{brown2020language}
Tom~B. Brown, Benjamin Mann, Nick Ryder, Melanie Subbiah, Jared Kaplan, Prafulla Dhariwal, Arvind Neelakantan, Pranav Shyam, Girish Sastry, Amanda Askell, Sandhini Agarwal, Ariel Herbert-Voss, Gretchen Krueger, Tom Henighan, Rewon Child, Aditya Ramesh, Daniel~M. Ziegler, Jeffrey Wu, Clemens Winter, Christopher Hesse, Mark Chen, Eric Sigler, Mateusz Litwin, Scott Gray, Benjamin Chess, Jack Clark, Christopher Berner, Sam McCandlish, Alec Radford, Ilya Sutskever, and Dario Amodei.
\newblock Language models are few-shot learners, 2020.

\bibitem[Chen et~al.(2020)Chen, Frankle, Chang, Liu, Zhang, Wang, and Carbin]{chen2020lottery}
Tianlong Chen, Jonathan Frankle, Shiyu Chang, Sijia Liu, Yang Zhang, Zhangyang Wang, and Michael Carbin.
\newblock The lottery ticket hypothesis for pre-trained bert networks.
\newblock In H.~Larochelle, M.~Ranzato, R.~Hadsell, M.F. Balcan, and H.~Lin, editors, \emph{Advances in Neural Information Processing Systems}, volume~33, pages 15834--15846. Curran Associates, Inc., 2020.
\newblock URL \url{https://proceedings.neurips.cc/paper_files/paper/2020/file/b6af2c9703f203a2794be03d443af2e3-Paper.pdf}.

\bibitem[Chen et~al.(2021)Chen, Cheng, Wang, Gan, Wang, and Liu]{chen2021earlyBERT}
Xiaohan Chen, Yu~Cheng, Shuohang Wang, Zhe Gan, Zhangyang Wang, and Jingjing Liu.
\newblock Earlybert: Efficient bert training via early-bird lottery tickets, 2021.

\bibitem[Chmiel et~al.(2023)Chmiel, Hubara, Banner, and Soudry]{chmiel2023minimum}
Brian Chmiel, Itay Hubara, Ron Banner, and Daniel Soudry.
\newblock Minimum variance unbiased n:m sparsity for the neural gradients.
\newblock In \emph{The Eleventh International Conference on Learning Representations}, 2023.
\newblock URL \url{https://openreview.net/forum?id=vuD2xEtxZcj}.

\bibitem[Dathathri et~al.(2020)Dathathri, Madotto, Lan, Hung, Frank, Molino, Yosinski, and Liu]{dathathri2020plug}
Sumanth Dathathri, Andrea Madotto, Janice Lan, Jane Hung, Eric Frank, Piero Molino, Jason Yosinski, and Rosanne Liu.
\newblock Plug and play language models: A simple approach to controlled text generation, 2020.

\bibitem[Deng et~al.(2009)Deng, Dong, Socher, Li, Li, and Fei-Fei]{5206848}
Jia Deng, Wei Dong, Richard Socher, Li-Jia Li, Kai Li, and Li~Fei-Fei.
\newblock Imagenet: A large-scale hierarchical image database.
\newblock In \emph{2009 IEEE Conference on Computer Vision and Pattern Recognition}, pages 248--255, 2009.
\newblock \doi{10.1109/CVPR.2009.5206848}.

\bibitem[Evci et~al.(2020)Evci, Gale, Menick, Castro, and Elsen]{evci2021rigging}
Utku Evci, Trevor Gale, Jacob Menick, Pablo~Samuel Castro, and Erich Elsen.
\newblock Rigging the lottery: Making all tickets winners.
\newblock In Hal~Daumé III and Aarti Singh, editors, \emph{Proceedings of the 37th International Conference on Machine Learning}, volume 119 of \emph{Proceedings of Machine Learning Research}, pages 2943--2952. PMLR, 13--18 Jul 2020.
\newblock URL \url{https://proceedings.mlr.press/v119/evci20a.html}.

\bibitem[Frankle and Carbin(2019)]{frankle2018the}
Jonathan Frankle and Michael Carbin.
\newblock The lottery ticket hypothesis: Finding sparse, trainable neural networks.
\newblock In \emph{International Conference on Learning Representations}, 2019.
\newblock URL \url{https://openreview.net/forum?id=rJl-b3RcF7}.

\bibitem[Frankle et~al.(2020{\natexlab{a}})Frankle, Dziugaite, Roy, and Carbin]{frankle2020linear}
Jonathan Frankle, Gintare~Karolina Dziugaite, Daniel Roy, and Michael Carbin.
\newblock Linear mode connectivity and the lottery ticket hypothesis.
\newblock In Hal~Daumé III and Aarti Singh, editors, \emph{Proceedings of the 37th International Conference on Machine Learning}, volume 119 of \emph{Proceedings of Machine Learning Research}, pages 3259--3269. PMLR, 13--18 Jul 2020{\natexlab{a}}.
\newblock URL \url{https://proceedings.mlr.press/v119/frankle20a.html}.

\bibitem[Frankle et~al.(2020{\natexlab{b}})Frankle, Dziugaite, Roy, and Carbin]{frankle2020stabilizing}
Jonathan Frankle, Gintare~Karolina Dziugaite, Daniel~M. Roy, and Michael Carbin.
\newblock Stabilizing the lottery ticket hypothesis, 2020{\natexlab{b}}.

\bibitem[Frantar and Alistarh(2023)]{frantar2023sparsegpt}
Elias Frantar and Dan Alistarh.
\newblock Sparsegpt: Massive language models can be accurately pruned in one-shot, 2023.

\bibitem[Gokaslan and Cohen(2019)]{Gokaslan2019OpenWeb}
Aaron Gokaslan and Vanya Cohen.
\newblock Openwebtext corpus.
\newblock \url{http://Skylion007.github.io/OpenWebTextCorpus}, 2019.

\bibitem[Han et~al.(2015)Han, Pool, Tran, and Dally]{han2015learning}
Song Han, Jeff Pool, John Tran, and William~J. Dally.
\newblock Learning both weights and connections for efficient neural networks, 2015.

\bibitem[Han et~al.(2016)Han, Mao, and Dally]{han2016deep}
Song Han, Huizi Mao, and William~J. Dally.
\newblock Deep compression: Compressing deep neural networks with pruning, trained quantization and huffman coding, 2016.

\bibitem[He et~al.(2017)He, Zhang, and Sun]{he2017channel}
Yihui He, Xiangyu Zhang, and Jian Sun.
\newblock Channel pruning for accelerating very deep neural networks.
\newblock In \emph{The IEEE International Conference on Computer Vision (ICCV)}, Oct 2017.

\bibitem[Hu et~al.(2024)Hu, Zhao, Huang, Chen, and Zhu]{hu2024accelerating}
Yuezhou Hu, Kang Zhao, Weiyu Huang, Jianfei Chen, and Jun Zhu.
\newblock Accelerating transformer pre-training with 2:4 sparsity.
\newblock In Ruslan Salakhutdinov, Zico Kolter, Katherine Heller, Adrian Weller, Nuria Oliver, Jonathan Scarlett, and Felix Berkenkamp, editors, \emph{Proceedings of the 41st International Conference on Machine Learning}, volume 235 of \emph{Proceedings of Machine Learning Research}, pages 19531--19543. PMLR, 21--27 Jul 2024.
\newblock URL \url{https://proceedings.mlr.press/v235/hu24r.html}.

\bibitem[Hubara et~al.(2021)Hubara, Chmiel, Island, Banner, Naor, and Soudry]{hubara2021accelerated}
Itay Hubara, Brian Chmiel, Moshe Island, Ron Banner, Joseph Naor, and Daniel Soudry.
\newblock Accelerated sparse neural training: A provable and efficient method to find n:m transposable masks.
\newblock In M.~Ranzato, A.~Beygelzimer, Y.~Dauphin, P.S. Liang, and J.~Wortman Vaughan, editors, \emph{Advances in Neural Information Processing Systems}, volume~34, pages 21099--21111. Curran Associates, Inc., 2021.
\newblock URL \url{https://proceedings.neurips.cc/paper_files/paper/2021/file/b0490b85e92b64dbb5db76bf8fca6a82-Paper.pdf}.

\bibitem[Lagunas et~al.(2021)Lagunas, Charlaix, Sanh, and Rush]{lagunas2021block}
Fran{\c{c}}ois Lagunas, Ella Charlaix, Victor Sanh, and Alexander Rush.
\newblock Block pruning for faster transformers.
\newblock In Marie-Francine Moens, Xuanjing Huang, Lucia Specia, and Scott Wen-tau Yih, editors, \emph{Proceedings of the 2021 Conference on Empirical Methods in Natural Language Processing}, pages 10619--10629, Online and Punta Cana, Dominican Republic, November 2021. Association for Computational Linguistics.
\newblock \doi{10.18653/v1/2021.emnlp-main.829}.
\newblock URL \url{https://aclanthology.org/2021.emnlp-main.829}.

\bibitem[Lasby et~al.(2023)Lasby, Golubeva, Evci, Nica, and Ioannou]{lasby2023dynamic}
Mike Lasby, Anna Golubeva, Utku Evci, Mihai Nica, and Yani Ioannou.
\newblock Dynamic sparse training with structured sparsity, 2023.

\bibitem[Lee et~al.(2019)Lee, Ajanthan, and Torr]{lee2019snip}
Namhoon Lee, Thalaiyasingam Ajanthan, and Philip Torr.
\newblock {SNIP}: {SINGLE}-{SHOT} {NETWORK} {PRUNING} {BASED} {ON} {CONNECTION} {SENSITIVITY}.
\newblock In \emph{International Conference on Learning Representations}, 2019.
\newblock URL \url{https://openreview.net/forum?id=B1VZqjAcYX}.

\bibitem[Li and Yuan(2017)]{li2017convergence}
Yuanzhi Li and Yang Yuan.
\newblock Convergence analysis of two-layer neural networks with relu activation.
\newblock In I.~Guyon, U.~Von Luxburg, S.~Bengio, H.~Wallach, R.~Fergus, S.~Vishwanathan, and R.~Garnett, editors, \emph{Advances in Neural Information Processing Systems}, volume~30. Curran Associates, Inc., 2017.
\newblock URL \url{https://proceedings.neurips.cc/paper_files/paper/2017/file/a96b65a721e561e1e3de768ac819ffbb-Paper.pdf}.

\bibitem[Lin et~al.(2020)Lin, Han, Mao, Wang, and Dally]{lin2020deep}
Yujun Lin, Song Han, Huizi Mao, Yu~Wang, and William~J. Dally.
\newblock Deep gradient compression: Reducing the communication bandwidth for distributed training, 2020.

\bibitem[Liu et~al.(2024)Liu, Zhang, Li, Yan, Gao, Chen, Yuan, Huang, Sun, Gao, He, and Sun]{liu2024sora}
Yixin Liu, Kai Zhang, Yuan Li, Zhiling Yan, Chujie Gao, Ruoxi Chen, Zhengqing Yuan, Yue Huang, Hanchi Sun, Jianfeng Gao, Lifang He, and Lichao Sun.
\newblock Sora: A review on background, technology, limitations, and opportunities of large vision models, 2024.

\bibitem[Lu et~al.(2023)Lu, Agrawal, Subramanian, Rybakov, De~Sa, and Yazdanbakhsh]{lu2023step}
Yucheng Lu, Shivani Agrawal, Suvinay Subramanian, Oleg Rybakov, Christopher De~Sa, and Amir Yazdanbakhsh.
\newblock {STEP}: Learning {N}:{M} structured sparsity masks from scratch with precondition.
\newblock In Andreas Krause, Emma Brunskill, Kyunghyun Cho, Barbara Engelhardt, Sivan Sabato, and Jonathan Scarlett, editors, \emph{Proceedings of the 40th International Conference on Machine Learning}, volume 202 of \emph{Proceedings of Machine Learning Research}, pages 22812--22824. PMLR, 23--29 Jul 2023.
\newblock URL \url{https://proceedings.mlr.press/v202/lu23c.html}.

\bibitem[Maene et~al.(2021)Maene, Li, and Moens]{maene2021understanding}
Jaron Maene, Mingxiao Li, and Marie-Francine Moens.
\newblock Towards understanding iterative magnitude pruning: Why lottery tickets win, 2021.

\bibitem[Micikevicius et~al.(2022)Micikevicius, Stosic, Burgess, Cornea, Dubey, Grisenthwaite, Ha, Heinecke, Judd, Kamalu, Mellempudi, Oberman, Shoeybi, Siu, and Wu]{micikevicius2022fp8}
Paulius Micikevicius, Dusan Stosic, Neil Burgess, Marius Cornea, Pradeep Dubey, Richard Grisenthwaite, Sangwon Ha, Alexander Heinecke, Patrick Judd, John Kamalu, Naveen Mellempudi, Stuart Oberman, Mohammad Shoeybi, Michael Siu, and Hao Wu.
\newblock Fp8 formats for deep learning, 2022.

\bibitem[Mishra et~al.(2021)Mishra, Latorre, Pool, Stosic, Stosic, Venkatesh, Yu, and Micikevicius]{mishra2021accelerating}
Asit Mishra, Jorge~Albericio Latorre, Jeff Pool, Darko Stosic, Dusan Stosic, Ganesh Venkatesh, Chong Yu, and Paulius Micikevicius.
\newblock Accelerating sparse deep neural networks, 2021.

\bibitem[Noune et~al.(2022)Noune, Jones, Justus, Masters, and Luschi]{noune20228bit}
Badreddine Noune, Philip Jones, Daniel Justus, Dominic Masters, and Carlo Luschi.
\newblock 8-bit numerical formats for deep neural networks, 2022.

\bibitem[OpenAI et~al.(2024)OpenAI, Achiam, Adler, Agarwal, Ahmad, Akkaya, Aleman, Almeida, Altenschmidt, Altman, Anadkat, Avila, Babuschkin, Balaji, Balcom, Baltescu, Bao, Bavarian, Belgum, Bello, Berdine, Bernadett-Shapiro, Berner, Bogdonoff, Boiko, Boyd, Brakman, Brockman, Brooks, Brundage, Button, Cai, Campbell, Cann, Carey, Carlson, Carmichael, Chan, Chang, Chantzis, Chen, Chen, Chen, Chen, Chen, Chess, Cho, Chu, Chung, Cummings, Currier, Dai, Decareaux, Degry, Deutsch, Deville, Dhar, Dohan, Dowling, Dunning, Ecoffet, Eleti, Eloundou, Farhi, Fedus, Felix, Fishman, Forte, Fulford, Gao, Georges, Gibson, Goel, Gogineni, Goh, Gontijo-Lopes, Gordon, Grafstein, Gray, Greene, Gross, Gu, Guo, Hallacy, Han, Harris, He, Heaton, Heidecke, Hesse, Hickey, Hickey, Hoeschele, Houghton, Hsu, Hu, Hu, Huizinga, Jain, Jain, Jang, Jiang, Jiang, Jin, Jin, Jomoto, Jonn, Jun, Kaftan, Łukasz Kaiser, Kamali, Kanitscheider, Keskar, Khan, Kilpatrick, Kim, Kim, Kim, Kirchner, Kiros, Knight, Kokotajlo, Łukasz Kondraciuk, Kondrich,
  Konstantinidis, Kosic, Krueger, Kuo, Lampe, Lan, Lee, Leike, Leung, Levy, Li, Lim, Lin, Lin, Litwin, Lopez, Lowe, Lue, Makanju, Malfacini, Manning, Markov, Markovski, Martin, Mayer, Mayne, McGrew, McKinney, McLeavey, McMillan, McNeil, Medina, Mehta, Menick, Metz, Mishchenko, Mishkin, Monaco, Morikawa, Mossing, Mu, Murati, Murk, Mély, Nair, Nakano, Nayak, Neelakantan, Ngo, Noh, Ouyang, O'Keefe, Pachocki, Paino, Palermo, Pantuliano, Parascandolo, Parish, Parparita, Passos, Pavlov, Peng, Perelman, de~Avila Belbute~Peres, Petrov, de~Oliveira~Pinto, Michael, Pokorny, Pokrass, Pong, Powell, Power, Power, Proehl, Puri, Radford, Rae, Ramesh, Raymond, Real, Rimbach, Ross, Rotsted, Roussez, Ryder, Saltarelli, Sanders, Santurkar, Sastry, Schmidt, Schnurr, Schulman, Selsam, Sheppard, Sherbakov, Shieh, Shoker, Shyam, Sidor, Sigler, Simens, Sitkin, Slama, Sohl, Sokolowsky, Song, Staudacher, Such, Summers, Sutskever, Tang, Tezak, Thompson, Tillet, Tootoonchian, Tseng, Tuggle, Turley, Tworek, Uribe, Vallone, Vijayvergiya,
  Voss, Wainwright, Wang, Wang, Wang, Ward, Wei, Weinmann, Welihinda, Welinder, Weng, Weng, Wiethoff, Willner, Winter, Wolrich, Wong, Workman, Wu, Wu, Wu, Xiao, Xu, Yoo, Yu, Yuan, Zaremba, Zellers, Zhang, Zhang, Zhao, Zheng, Zhuang, Zhuk, and Zoph]{openai2024gpt4}
OpenAI, Josh Achiam, Steven Adler, Sandhini Agarwal, Lama Ahmad, Ilge Akkaya, Florencia~Leoni Aleman, Diogo Almeida, Janko Altenschmidt, Sam Altman, Shyamal Anadkat, Red Avila, Igor Babuschkin, Suchir Balaji, Valerie Balcom, Paul Baltescu, Haiming Bao, Mohammad Bavarian, Jeff Belgum, Irwan Bello, Jake Berdine, Gabriel Bernadett-Shapiro, Christopher Berner, Lenny Bogdonoff, Oleg Boiko, Madelaine Boyd, Anna-Luisa Brakman, Greg Brockman, Tim Brooks, Miles Brundage, Kevin Button, Trevor Cai, Rosie Campbell, Andrew Cann, Brittany Carey, Chelsea Carlson, Rory Carmichael, Brooke Chan, Che Chang, Fotis Chantzis, Derek Chen, Sully Chen, Ruby Chen, Jason Chen, Mark Chen, Ben Chess, Chester Cho, Casey Chu, Hyung~Won Chung, Dave Cummings, Jeremiah Currier, Yunxing Dai, Cory Decareaux, Thomas Degry, Noah Deutsch, Damien Deville, Arka Dhar, David Dohan, Steve Dowling, Sheila Dunning, Adrien Ecoffet, Atty Eleti, Tyna Eloundou, David Farhi, Liam Fedus, Niko Felix, Simón~Posada Fishman, Juston Forte, Isabella Fulford, Leo
  Gao, Elie Georges, Christian Gibson, Vik Goel, Tarun Gogineni, Gabriel Goh, Rapha Gontijo-Lopes, Jonathan Gordon, Morgan Grafstein, Scott Gray, Ryan Greene, Joshua Gross, Shixiang~Shane Gu, Yufei Guo, Chris Hallacy, Jesse Han, Jeff Harris, Yuchen He, Mike Heaton, Johannes Heidecke, Chris Hesse, Alan Hickey, Wade Hickey, Peter Hoeschele, Brandon Houghton, Kenny Hsu, Shengli Hu, Xin Hu, Joost Huizinga, Shantanu Jain, Shawn Jain, Joanne Jang, Angela Jiang, Roger Jiang, Haozhun Jin, Denny Jin, Shino Jomoto, Billie Jonn, Heewoo Jun, Tomer Kaftan, Łukasz Kaiser, Ali Kamali, Ingmar Kanitscheider, Nitish~Shirish Keskar, Tabarak Khan, Logan Kilpatrick, Jong~Wook Kim, Christina Kim, Yongjik Kim, Jan~Hendrik Kirchner, Jamie Kiros, Matt Knight, Daniel Kokotajlo, Łukasz Kondraciuk, Andrew Kondrich, Aris Konstantinidis, Kyle Kosic, Gretchen Krueger, Vishal Kuo, Michael Lampe, Ikai Lan, Teddy Lee, Jan Leike, Jade Leung, Daniel Levy, Chak~Ming Li, Rachel Lim, Molly Lin, Stephanie Lin, Mateusz Litwin, Theresa Lopez, Ryan
  Lowe, Patricia Lue, Anna Makanju, Kim Malfacini, Sam Manning, Todor Markov, Yaniv Markovski, Bianca Martin, Katie Mayer, Andrew Mayne, Bob McGrew, Scott~Mayer McKinney, Christine McLeavey, Paul McMillan, Jake McNeil, David Medina, Aalok Mehta, Jacob Menick, Luke Metz, Andrey Mishchenko, Pamela Mishkin, Vinnie Monaco, Evan Morikawa, Daniel Mossing, Tong Mu, Mira Murati, Oleg Murk, David Mély, Ashvin Nair, Reiichiro Nakano, Rajeev Nayak, Arvind Neelakantan, Richard Ngo, Hyeonwoo Noh, Long Ouyang, Cullen O'Keefe, Jakub Pachocki, Alex Paino, Joe Palermo, Ashley Pantuliano, Giambattista Parascandolo, Joel Parish, Emy Parparita, Alex Passos, Mikhail Pavlov, Andrew Peng, Adam Perelman, Filipe de~Avila Belbute~Peres, Michael Petrov, Henrique~Ponde de~Oliveira~Pinto, Michael, Pokorny, Michelle Pokrass, Vitchyr~H. Pong, Tolly Powell, Alethea Power, Boris Power, Elizabeth Proehl, Raul Puri, Alec Radford, Jack Rae, Aditya Ramesh, Cameron Raymond, Francis Real, Kendra Rimbach, Carl Ross, Bob Rotsted, Henri Roussez,
  Nick Ryder, Mario Saltarelli, Ted Sanders, Shibani Santurkar, Girish Sastry, Heather Schmidt, David Schnurr, John Schulman, Daniel Selsam, Kyla Sheppard, Toki Sherbakov, Jessica Shieh, Sarah Shoker, Pranav Shyam, Szymon Sidor, Eric Sigler, Maddie Simens, Jordan Sitkin, Katarina Slama, Ian Sohl, Benjamin Sokolowsky, Yang Song, Natalie Staudacher, Felipe~Petroski Such, Natalie Summers, Ilya Sutskever, Jie Tang, Nikolas Tezak, Madeleine~B. Thompson, Phil Tillet, Amin Tootoonchian, Elizabeth Tseng, Preston Tuggle, Nick Turley, Jerry Tworek, Juan Felipe~Cerón Uribe, Andrea Vallone, Arun Vijayvergiya, Chelsea Voss, Carroll Wainwright, Justin~Jay Wang, Alvin Wang, Ben Wang, Jonathan Ward, Jason Wei, CJ~Weinmann, Akila Welihinda, Peter Welinder, Jiayi Weng, Lilian Weng, Matt Wiethoff, Dave Willner, Clemens Winter, Samuel Wolrich, Hannah Wong, Lauren Workman, Sherwin Wu, Jeff Wu, Michael Wu, Kai Xiao, Tao Xu, Sarah Yoo, Kevin Yu, Qiming Yuan, Wojciech Zaremba, Rowan Zellers, Chong Zhang, Marvin Zhang, Shengjia
  Zhao, Tianhao Zheng, Juntang Zhuang, William Zhuk, and Barret Zoph.
\newblock Gpt-4 technical report, 2024.

\bibitem[Ott et~al.(2019)Ott, Edunov, Baevski, Fan, Gross, Ng, Grangier, and Auli]{ott2019fairseq}
Myle Ott, Sergey Edunov, Alexei Baevski, Angela Fan, Sam Gross, Nathan Ng, David Grangier, and Michael Auli.
\newblock fairseq: A fast, extensible toolkit for sequence modeling.
\newblock In \emph{Proceedings of NAACL-HLT 2019: Demonstrations}, 2019.

\bibitem[Pandey et~al.(2007)Pandey, Delorey, Duan, Wang, Knutson, Zappala, and Woodings]{Pandey2007RIAAR}
Manoj~Kumar Pandey, Daniel~P. Delorey, Qiuyi Duan, Lei Wang, Charles~D. Knutson, Daniel Zappala, and Ryan Woodings.
\newblock Ria: An rf interference avoidance algorithm for heterogeneous wireless networks.
\newblock \emph{2007 IEEE Wireless Communications and Networking Conference}, pages 4051--4056, 2007.
\newblock URL \url{https://api.semanticscholar.org/CorpusID:10798336}.

\bibitem[Papineni et~al.(2002)Papineni, Roukos, Ward, and Zhu]{article}
Kishore Papineni, Salim Roukos, Todd Ward, and Wei-Jing Zhu.
\newblock {B}leu: a method for automatic evaluation of machine translation.
\newblock In Pierre Isabelle, Eugene Charniak, and Dekang Lin, editors, \emph{Proceedings of the 40th Annual Meeting of the Association for Computational Linguistics}, pages 311--318, Philadelphia, Pennsylvania, USA, July 2002. Association for Computational Linguistics.
\newblock \doi{10.3115/1073083.1073135}.
\newblock URL \url{https://aclanthology.org/P02-1040}.

\bibitem[Perez et~al.(2023)Perez, Zhang, Briggs, Blake, Levy-Kramer, Balanca, Luschi, Barlow, and Fitzgibbon]{perez2023training}
Sergio~P. Perez, Yan Zhang, James Briggs, Charlie Blake, Josh Levy-Kramer, Paul Balanca, Carlo Luschi, Stephen Barlow, and Andrew~William Fitzgibbon.
\newblock Training and inference of large language models using 8-bit floating point, 2023.

\bibitem[Radford et~al.(2019)Radford, Wu, Child, Luan, Amodei, and Sutskever]{Radford2019LanguageMA}
Alec Radford, Jeff Wu, Rewon Child, David Luan, Dario Amodei, and Ilya Sutskever.
\newblock Language models are unsupervised multitask learners.
\newblock 2019.

\bibitem[Rajpurkar et~al.(2016)Rajpurkar, Zhang, Lopyrev, and Liang]{rajpurkar2016squad}
Pranav Rajpurkar, Jian Zhang, Konstantin Lopyrev, and Percy Liang.
\newblock {SQ}u{AD}: 100,000+ questions for machine comprehension of text.
\newblock In Jian Su, Kevin Duh, and Xavier Carreras, editors, \emph{Proceedings of the 2016 Conference on Empirical Methods in Natural Language Processing}, pages 2383--2392, Austin, Texas, November 2016. Association for Computational Linguistics.
\newblock \doi{10.18653/v1/D16-1264}.
\newblock URL \url{https://aclanthology.org/D16-1264}.

\bibitem[Sun et~al.(2023)Sun, Liu, Bair, and Kolter]{sun2023wanda}
Mingjie Sun, Zhuang Liu, Anna Bair, and J.~Zico Kolter.
\newblock A simple and effective pruning approach for large language models.
\newblock \emph{arXiv preprint arXiv:2306.11695}, 2023.

\bibitem[Sun et~al.(2019)Sun, Choi, Chen, Wang, Venkataramani, Srinivasan, Cui, Zhang, and Gopalakrishnan]{Sun2019Hybrid8F}
Xiao Sun, Jungwook Choi, Chia-Yu Chen, Naigang Wang, Swagath Venkataramani, Vijayalakshmi Srinivasan, Xiaodong Cui, Wei Zhang, and K.~Gopalakrishnan.
\newblock Hybrid 8-bit floating point (hfp8) training and inference for deep neural networks.
\newblock In \emph{Neural Information Processing Systems}, 2019.
\newblock URL \url{https://api.semanticscholar.org/CorpusID:202779157}.

\bibitem[Touvron et~al.(2021)Touvron, Cord, Douze, Massa, Sablayrolles, and Jégou]{touvron2021training}
Hugo Touvron, Matthieu Cord, Matthijs Douze, Francisco Massa, Alexandre Sablayrolles, and Hervé Jégou.
\newblock Training data-efficient image transformers \& distillation through attention, 2021.

\bibitem[Touvron et~al.(2023)Touvron, Lavril, Izacard, Martinet, Lachaux, Lacroix, Rozi{\`e}re, Goyal, Hambro, Azhar, et~al.]{touvron2023llama}
Hugo Touvron, Thibaut Lavril, Gautier Izacard, Xavier Martinet, Marie-Anne Lachaux, Timoth{\'e}e Lacroix, Baptiste Rozi{\`e}re, Naman Goyal, Eric Hambro, Faisal Azhar, et~al.
\newblock Llama: Open and efficient foundation language models.
\newblock \emph{arXiv preprint arXiv:2302.13971}, 2023.

\bibitem[van Baalen et~al.(2023)van Baalen, Kuzmin, Nair, Ren, Mahurin, Patel, Subramanian, Lee, Nagel, Soriaga, and Blankevoort]{vanbaalen2023fp8}
Mart van Baalen, Andrey Kuzmin, Suparna~S Nair, Yuwei Ren, Eric Mahurin, Chirag Patel, Sundar Subramanian, Sanghyuk Lee, Markus Nagel, Joseph Soriaga, and Tijmen Blankevoort.
\newblock Fp8 versus int8 for efficient deep learning inference, 2023.

\bibitem[Vanderschueren and Vleeschouwer(2022)]{vanderschueren2022straightthrough}
Antoine Vanderschueren and Christophe~De Vleeschouwer.
\newblock Are straight-through gradients and soft-thresholding all you need for sparse training?, 2022.

\bibitem[Vaswani et~al.(2017)Vaswani, Shazeer, Parmar, Uszkoreit, Jones, Gomez, Kaiser, and Polosukhin]{vaswani2023attention}
Ashish Vaswani, Noam Shazeer, Niki Parmar, Jakob Uszkoreit, Llion Jones, Aidan~N Gomez, \L~ukasz Kaiser, and Illia Polosukhin.
\newblock Attention is all you need.
\newblock In I.~Guyon, U.~Von Luxburg, S.~Bengio, H.~Wallach, R.~Fergus, S.~Vishwanathan, and R.~Garnett, editors, \emph{Advances in Neural Information Processing Systems}, volume~30. Curran Associates, Inc., 2017.
\newblock URL \url{https://proceedings.neurips.cc/paper_files/paper/2017/file/3f5ee243547dee91fbd053c1c4a845aa-Paper.pdf}.

\bibitem[Wang et~al.(2018{\natexlab{a}})Wang, Singh, Michael, Hill, Levy, and Bowman]{wang2019glue}
Alex Wang, Amanpreet Singh, Julian Michael, Felix Hill, Omer Levy, and Samuel Bowman.
\newblock {GLUE}: A multi-task benchmark and analysis platform for natural language understanding.
\newblock In Tal Linzen, Grzegorz Chrupa{\l}a, and Afra Alishahi, editors, \emph{Proceedings of the 2018 {EMNLP} Workshop {B}lackbox{NLP}: Analyzing and Interpreting Neural Networks for {NLP}}, pages 353--355, Brussels, Belgium, November 2018{\natexlab{a}}. Association for Computational Linguistics.
\newblock \doi{10.18653/v1/W18-5446}.
\newblock URL \url{https://aclanthology.org/W18-5446}.

\bibitem[Wang et~al.(2018{\natexlab{b}})Wang, Choi, Brand, Chen, and Gopalakrishnan]{wang2018training}
Naigang Wang, Jungwook Choi, Daniel Brand, Chia-Yu Chen, and Kailash Gopalakrishnan.
\newblock Training deep neural networks with 8-bit floating point numbers, 2018{\natexlab{b}}.

\bibitem[Yin et~al.(2023)Yin, Li, Fang, Shen, Huang, Wang, Menkovski, Ma, Pechenizkiy, and Liu]{yin2023dynamic}
Lu~Yin, Gen Li, Meng Fang, Li~Shen, Tianjin Huang, Zhangyang~"Atlas" Wang, Vlado Menkovski, Xiaolong Ma, Mykola Pechenizkiy, and Shiwei Liu.
\newblock Dynamic sparsity is channel-level sparsity learner.
\newblock In A.~Oh, T.~Naumann, A.~Globerson, K.~Saenko, M.~Hardt, and S.~Levine, editors, \emph{Advances in Neural Information Processing Systems}, volume~36, pages 67993--68012. Curran Associates, Inc., 2023.
\newblock URL \url{https://proceedings.neurips.cc/paper_files/paper/2023/file/d6d0e41e0b1ed38c76d13c9e417a8f1f-Paper-Conference.pdf}.

\bibitem[You et~al.(2022)You, Li, Xu, Fu, Wang, Chen, Baraniuk, Wang, and Lin]{you2022drawing}
Haoran You, Chaojian Li, Pengfei Xu, Yonggan Fu, Yue Wang, Xiaohan Chen, Richard~G. Baraniuk, Zhangyang Wang, and Yingyan Lin.
\newblock Drawing early-bird tickets: Towards more efficient training of deep networks, 2022.

\bibitem[Zhang et~al.(2023)Zhang, Luo, Lin, Zhong, Xie, Chao, and Ji]{zhang2023bidirectional}
Yuxin Zhang, Yiting Luo, Mingbao Lin, Yunshan Zhong, Jingjing Xie, Fei Chao, and Rongrong Ji.
\newblock Bi-directional masks for efficient {N}:{M} sparse training.
\newblock In Andreas Krause, Emma Brunskill, Kyunghyun Cho, Barbara Engelhardt, Sivan Sabato, and Jonathan Scarlett, editors, \emph{Proceedings of the 40th International Conference on Machine Learning}, volume 202 of \emph{Proceedings of Machine Learning Research}, pages 41488--41497. PMLR, 23--29 Jul 2023.
\newblock URL \url{https://proceedings.mlr.press/v202/zhang23ae.html}.

\bibitem[Zhou et~al.(2021)Zhou, Ma, Zhu, Liu, Zhang, Yuan, Sun, and Li]{zhou2021learning}
Aojun Zhou, Yukun Ma, Junnan Zhu, Jianbo Liu, Zhijie Zhang, Kun Yuan, Wenxiu Sun, and Hongsheng Li.
\newblock Learning n:m fine-grained structured sparse neural networks from scratch.
\newblock In \emph{International Conference on Learning Representations}, 2021.
\newblock URL \url{https://openreview.net/forum?id=K9bw7vqp_s}.

\bibitem[Zitkovich et~al.(2023)Zitkovich, Yu, Xu, Xu, Xiao, Xia, Wu, Wohlhart, Welker, Wahid, Vuong, Vanhoucke, Tran, Soricut, Singh, Singh, Sermanet, Sanketi, Salazar, Ryoo, Reymann, Rao, Pertsch, Mordatch, Michalewski, Lu, Levine, Lee, Lee, Leal, Kuang, Kalashnikov, Julian, Joshi, Irpan, Ichter, Hsu, Herzog, Hausman, Gopalakrishnan, Fu, Florence, Finn, Dubey, Driess, Ding, Choromanski, Chen, Chebotar, Carbajal, Brown, Brohan, Arenas, and Han]{brohan2023rt2}
Brianna Zitkovich, Tianhe Yu, Sichun Xu, Peng Xu, Ted Xiao, Fei Xia, Jialin Wu, Paul Wohlhart, Stefan Welker, Ayzaan Wahid, Quan Vuong, Vincent Vanhoucke, Huong Tran, Radu Soricut, Anikait Singh, Jaspiar Singh, Pierre Sermanet, Pannag~R. Sanketi, Grecia Salazar, Michael~S. Ryoo, Krista Reymann, Kanishka Rao, Karl Pertsch, Igor Mordatch, Henryk Michalewski, Yao Lu, Sergey Levine, Lisa Lee, Tsang-Wei~Edward Lee, Isabel Leal, Yuheng Kuang, Dmitry Kalashnikov, Ryan Julian, Nikhil~J. Joshi, Alex Irpan, Brian Ichter, Jasmine Hsu, Alexander Herzog, Karol Hausman, Keerthana Gopalakrishnan, Chuyuan Fu, Pete Florence, Chelsea Finn, Kumar~Avinava Dubey, Danny Driess, Tianli Ding, Krzysztof~Marcin Choromanski, Xi~Chen, Yevgen Chebotar, Justice Carbajal, Noah Brown, Anthony Brohan, Montserrat~Gonzalez Arenas, and Kehang Han.
\newblock Rt-2: Vision-language-action models transfer web knowledge to robotic control.
\newblock In Jie Tan, Marc Toussaint, and Kourosh Darvish, editors, \emph{Proceedings of The 7th Conference on Robot Learning}, volume 229 of \emph{Proceedings of Machine Learning Research}, pages 2165--2183. PMLR, 06--09 Nov 2023.
\newblock URL \url{https://proceedings.mlr.press/v229/zitkovich23a.html}.

\end{thebibliography}
